\documentclass[sn-mathphys-num]{sn-jnl}

\usepackage{graphicx}%
\usepackage{multirow}%
\usepackage{amsmath,amssymb,amsfonts}%
\usepackage{amsthm}%
\usepackage{mathrsfs}%
\usepackage[title]{appendix}%
\usepackage{xcolor}%
\usepackage{textcomp}%
\usepackage{manyfoot}%
\usepackage{booktabs}%
\usepackage[ruled]{algorithm2e}
\usepackage{algpseudocode}%
\usepackage{listings}%
\usepackage{bm}
\usepackage{float}
\usepackage{tabularx}
\usepackage{makecell}
\usepackage[font={small}]{caption}

\theoremstyle{thmstyleone}%

\theoremstyle{thmstyletwo}%

\theoremstyle{thmstylethree}%

\makeatletter
\DeclareRobustCommand*\uell{\mathpalette\@uell\relax}
\newcommand*\@uell[2]{
  \setbox0=\hbox{$#1\ell$}
  \setbox1=\hbox{\rotatebox{10}{$#1\ell$}}
  \dimen0=\wd0 \advance\dimen0 by -\wd1 \divide\dimen0 by 2
  \mathord{\lower 0.3ex \hbox{\kern\dimen0\unhbox1\kern\dimen0}}
}

\DeclareMathOperator*{\argmax}{argmax}
\DeclareMathOperator*{\argmin}{argmin}

\begin{document}

\title[Batch-in-Batch: a new adversarial training framework for initial perturbation and sample selection]{Batch-in-Batch: a new adversarial training framework for initial perturbation and sample selection}

\author[1]{\fnm{Yinting} \sur{Wu}}

\author[2]{\fnm{Pai} \sur{Peng}}

\author[3]{\fnm{Bo} \sur{Cai}}

\author*[1]{\fnm{Le} \sur{Li}}\email{leli@mail.ccnu.edu.cn}

\affil[1]{
    \orgdiv{
        School of Mathematics and Statistics, and Key Lab NAA--MOE},
    \orgname{Central China Normal University},
    \orgaddress{
        \city{Wuhan},
        \country{China}}}

\affil[2]{
    \orgdiv{School of Mathematics and Computer Science}, 
    \orgname{Jianghan University}, 
    \orgaddress{
        \city{Wuhan}, 
        \country{China}}}

\affil[3]{
    \orgdiv{Key Laboratory of Aerospace Information Security and Trusted  Computing, Ministry of Education, and School of Cyber Science and Engineering}, 
    \orgname{Wuhan University}, 
    \orgaddress{
        \country{China}}}\vspace{-0.3em}


\abstract{Adversarial training methods commonly generate independent initial perturbation for adversarial samples from a simple uniform distribution, and obtain the training batch for the classifier without selection. In this work, we propose a simple yet effective training framework called Batch-in-Batch (BB) to enhance models robustness. It involves specifically a joint construction of initial values that could simultaneously generates \textit{m} sets of perturbations from the original batch set to provide more diversity for adversarial samples; and also includes various sample selection strategies that enable the trained models to have smoother losses and avoid overconfident outputs. Through extensive experiments on three benchmark datasets (CIFAR-10, SVHN, CIFAR-100) with two networks (PreActResNet18 and WideResNet28-10) that are used in both the single-step (Noise-Fast Gradient Sign Method, N-FGSM) and multi-step (Projected Gradient Descent, PGD-10) adversarial training, we show that models trained within the BB framework consistently have higher adversarial accuracy across various adversarial settings, notably achieving over a 13\% improvement on the SVHN dataset with an attack radius of 8/255 compared to the N-FGSM baseline model. Furthermore, experimental analysis of the efficiency of both the proposed initial perturbation method and sample selection strategies validates our insights. Finally, we show that our framework is cost-effective in terms of computational resources, even with a relatively large value of \textit{m}.}


\maketitle

\section{Introduction}\label{sec1}

By training \textit{deep neural networks} (DNNs) with maliciously perturbed samples, \textit{adversarial training} (AT) \cite{chen2020adversarial, goodfellow2014explaining, BIGGIO2018317, chakraborty2018adversarial} stands as one of the most effective methods to regularize a model to be robust against strong adversarial attacks \cite{madry2018towards, pang2020boosting}. Many works \cite{wong2020fast, andriushchenko2020understanding, madry2018towards, Pang2022RobustnessAA} have investigated how to perform AT efficiently and effectively by leveraging local gradient information.

Single-step AT aims at perturbing original samples with only one round of gradient computation. Goodfellow et al. \cite{goodfellow2014explaining} proposed \textit{fast gradient sign method} (FGSM), an adversarial attack prototype that generates adversarial samples by utilizing gradient information relative to the target loss on the original samples. 
To harness the problem that models trained with FGSM exhibit proficiency in generating degraded adversarial samples rather than acquiring robustness against adversarial samples, an approach called R+FGSM \cite{tramer2017ensemble} was proposed, where it inserts an initial perturbation step constrained by the sign function before performing FGSM. Later study \cite{wong2020fast} found that these single-step methods suffer from \textit{catastrophic overfitting} (CO), rendering adversarial accuracy against multi-step attack close to zero on the test set after certain epochs of training. Since then, many works focused to mitigate this problem while reserving the efficiency of single-step AT in the same time \cite{jorge2022make, ortiz-jimenez2023catastrophic, andriushchenko2020understanding, 10208721, ilyas2019adversarial}. Andriushchenko et al. \cite{andriushchenko2020understanding} established a connection between CO and the local linearity of loss function, and hence proposed GradAlign to regularize the model with a penalty term on the linearity of loss function. Meanwhile, some other works aim to address this problem with lower computational expenses by incorporating more effective initial perturbations. RS-FGSM \cite{wong2020fast} performs initial perturbation with noise drawn from an uniform distribution and eludes CO partially. N-FGSM \cite{jorge2022make} further chooses to discard the clip procedure and the $\uell_{\infty}$ restriction in previous methods, yielding comparable results to GradAlign, and three times faster than it.

Although originally intended to address overfitting, initial perturbation for adversarial sample also shows potential on improving model's robustness. For example, N-FGSM shows a similar ability to GradAlign in regularizing the training model to keep better local linearity \cite{jorge2022make}. But for now, initial perturbations used in most AT methods are independent across epochs, acquiescing the possibility of training on similar adversarial samples in consecutive epochs. To better exploit the prowess of initial perturbations, we argue that, by jointly constructing initial perturbations for multiple adversarial samples of one original sample, it is possible to fully utilize gradient information around the original sample and obtain better approximations of the local target loss.

In addition, Kim et al.\cite{kim2021understanding} proposed a method that prevents catastrophic overfitting by firstly inspecting several checkpoints on the adversarial direction of FGSM to revise its step length of perturbation, and then selects the one that induces an incorrect classification with minimum step length. This selection procedure of checkpoints can be interpreted as an indication that some adversarial samples in AT are unnecessary and may even be harmful. Another evidence supporting this point of view is the early stopping for AT \cite{rice2020overfitting}, which breaks the training process when the adversarial accuracy on the validation set ceases to increase for a certain number of epochs. Although the checkpoint method and early stopping control the training process at different levels, they can be generally regarded as training selectively on adversarial samples, meaning that those samples are not treated equally during training. As a result, it inspires us that a strategy dedicated to selecting adversarial samples during training may further boost model robustness.

\begin{figure}[t]
\hspace{-1.5cm}
\includegraphics[width=1.2\linewidth]{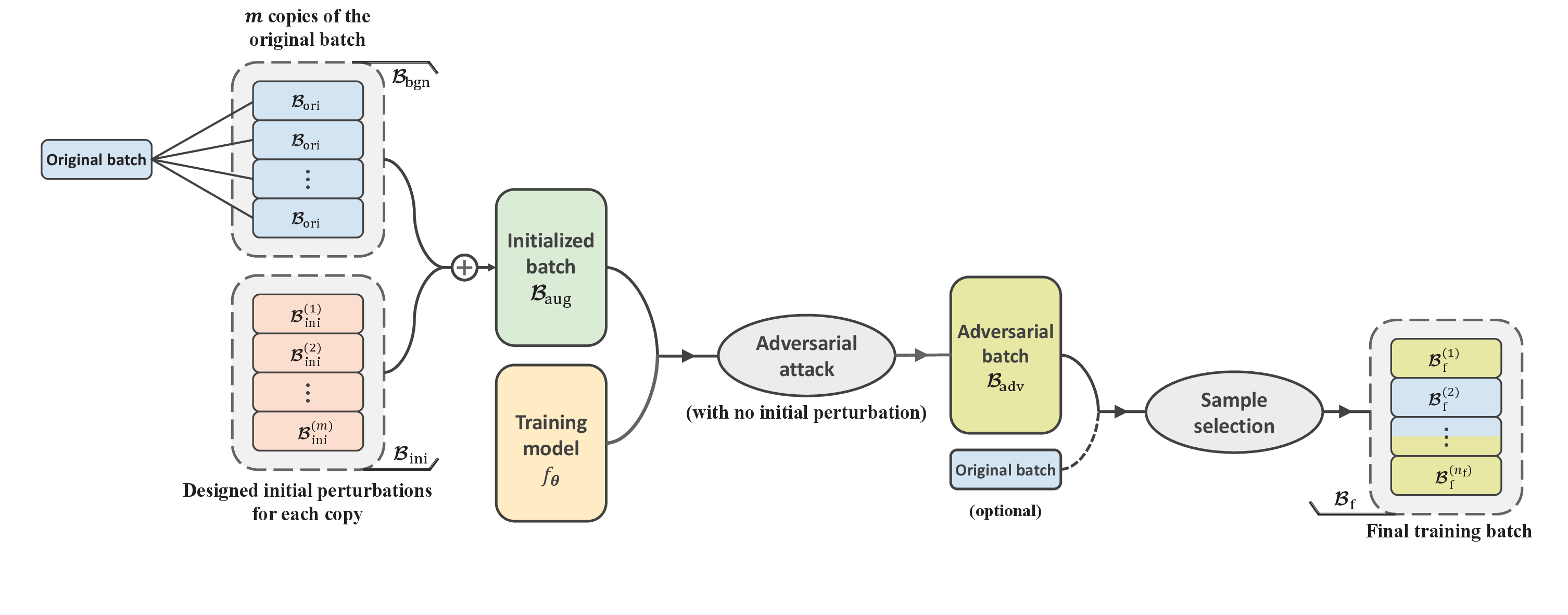}
\vspace{-0.8 cm}
\caption{The structure of Batch-in-Batch framework}
\vspace{-0.4cm}
\label{fig:overall}
\end{figure}

\begin{figure}[hbt!]
\vspace{-0.4cm}
\includegraphics[width=12em]{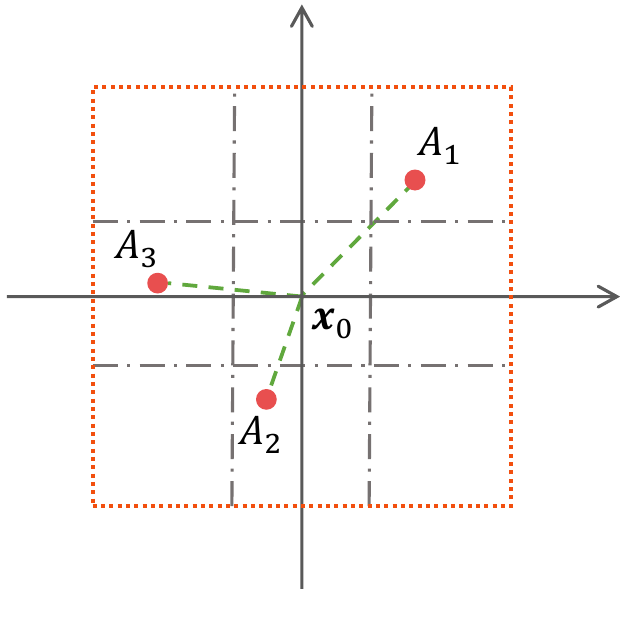}
\vspace{-0.1cm}
\centering
\caption{An 2D example of LHS. Given an original sample $\bm{x}_0$, the $\uell_{\infty}(\epsilon)$ vicinity and a desired sample number $m=3$, LHS firstly divides the vicinity with $m^2=9$ squares ($m^d$ hypercubes for $d>3$ dimensional spaces), then draws samples in orthogonally selected squares}
\label{fig:LHS_instance}
\end{figure}

We notice that the two ideas mentioned above can be achieved by generating multiple adversarial samples in parallel for each original sample. With this notion, we propose a simple yet effective training framework called Batch-in-Batch (BB), which serves not only for more delicate initial perturbations but also for adversarial sample selection (Figure \ref{fig:overall}). More precisely, given an original training batch $\mathcal{B}_{\mathrm{ori}}$, we firstly duplicate it $m$ times in parallel to form a larger batch $\mathcal{B}_{\mathrm{bgn}}$ for subsequent processing. Then, we employ a tuned \textit{Latin Hypercube Sampling} (LHS) \cite{LHS1979} to collectively design space-filling initial perturbations $\bm{\delta}_0$ for all copies in $\mathcal{B}_{\mathrm{bgn}}$. This approach aims to generate more separated initial perturbations from orthogonal hypercubes located around the original samples (see Figure \ref{fig:LHS_instance}). The initialized batch $\mathcal{B}_{\mathrm{aug}}$, obtained by perturbing $\mathcal{B}_{\mathrm{bgn}}$ with $\bm{\delta}_0$, is then processed by adversarial attacks and converted to an adversarial batch $\mathcal{B}_{\mathrm{adv}}$. Lastly, a sample selection strategy is conducted on $\mathcal{B}_{\mathrm{adv}}$ to form the final training batch $\mathcal{B}_{\mathrm{f}}$. One thing that needs to be clarified is that the initial perturbation, adversarial attack method, and selection procedure involved in the proposed framework can be arbitrary.

To evaluate the performance of proposed framework, we conduct experiments in multiple settings with various initial perturbations, adversarial attack methods, and sample selection procedures. For initial perturbations, apart from commonly used random initialization \cite{tramer2017ensemble, wong2020fast}, we propose a tuned LHS (tLHS) for it and conduct ablation experiments to estimate its effectiveness and computational cost; for adversarial attack methods, we consider N-FGSM and PGD \cite{madry2018towards} respectively as representative and baseline for single and multi-step adversarial attack; for sample selection strategies, we evaluate three crude methods: the first one is the same as Kim et al.\cite{kim2021understanding}, but with selecting target $\mathcal{B}_{\mathrm{adv}}$, instead of their original collinearly aligned checkpoints. The second one is an attempt to use early stopping on a datum-wise fashion, where we filter out all correctly classified samples in $\mathcal{B}_{\mathrm{adv}}$ to form $\mathcal{B}_{\mathrm{f}}$. The third one is a variant of the second, aiming to control the balance between clean and adversarial accuracy by performing sample selection on $\{\mathcal{B}_{\mathrm{adv}},\mathcal{B}_{\mathrm{ori}}\}$. Besides, we also provide empirical explanations supported by numerous experiments from various perspectives for the mechanisms of our framework.

The main contributions of this paper are as follows:
\begin{itemize}
  \item We propose a novel adversarial training framework for generating adversarial samples in parallel, which further enables not only conjunctive design of initial perturbation before an adversarial attack, but also sample selection after it.
  \item Under the proposed framework, we tested three sample selection strategies, two of which are original. The framework achieves decent improvements on adversarial accuracy against PGD-10 and N-FGSM over varied experimental settings.
  \item We delve into the explanation of the effectiveness of our framework in multiple perspectives, and provide experimental support.
\end{itemize}

\vspace{-1cm}

\section{Related work}\label{sec:related work}

\noindent \textbf{Random initialization of adversarial attacks}

Random initialization was first proposed by Tram{è}r et al. \cite{tramer2017ensemble} to reduce the likelihood of adversarial attacks being misled by a complex loss surface. They introduced random initialization into FGSM and proposed R+FGSM, where they draw each entry of the initial perturbations independently from a standard normal distribution, and project all of them onto $\{-1,1\}$ using a sign function.
However, the employment of the sign function in R+FGSM confines the elements of initial perturbations to a limited discrete set, thereby rendering the majority of points inaccessible to adversarial attacks.
Wong et al. \cite{wong2020fast} found this restricted perturbation initialization might be the cause of CO problem and proposed RS-FGSM to fix this issue by substituting initial perturbations drawn on an uniform distribution for R+FGSM initial perturbations. R+FGSM and RS-FGSM both restrict the $\uell _\infty$ norm of the final perturbation to be no greater than $\epsilon$ by either controlling the step length or performing a clip procedure. But still, RS-FGSM encroached by CO problem in AT training with larger $\epsilon$ \cite{jorge2022make}. de Jorge et al. \cite{jorge2022make} believe the restriction on the $\uell _\infty$ norm of adversarial perturbations impair the effectiveness of adversarial attack, since the initial perturbation consumed considerable budget of this restriction. They thus abandoned the restriction on the $\uell _\infty$ norm of adversarial perturbations, which also make it possible to initialize perturbations in a larger range than $\pm \epsilon$. Without the norm restriction, de Jorge et al. proposed N-FGSM, which augment initial perturbation radii $k$ times and successfully elude CO as a single-step AT method.

\vspace{0.2cm}
\noindent \textbf{Parallel generation}

There are few works that choose to generate multiple adversarial versions for one original sample in each step. Shafahi et al. \cite{shafahi2019adversarial} proposed Free-AT, whose main idea can be considered as performing FGSM repeatedly for one input batch with inherited perturbation from last round of FGSM. Free-AT generates adversarial samples one by one and updates model parameters using local gradient information after obtaining a new adversarial perturbation. Free-AT has successfully achieved comparable robustness to PGD and mostly kept the efficiency of single-step AT. Motivated by the relationship between CO and fixed step length of FGSM, Kim et al. \cite{kim2021understanding} brought up a new single-AT method in a checkpoint fashion, where they examine several evenly separated points on the adversarial direction provided by FGSM and select misclassified sample with minimum $\uell_{\infty}$ distance to the original sample. 

\section{Problem formulation}

For the ease of later discussion and experiments, we formulate the aim of adversarial attack and introduce two adversarial attack methods (\emph{i.e.,} N-FGSM and PGD) to be used in the experiments. Denote by $D=\{(\bm{x}_i,y_i) \ \vline\  \bm{x}_i \in \mathcal{X} \subset \mathbb{R}^d, y_i \in \mathcal{Y}, i=1,\ldots,N \}$ the sample set with feature space $\mathcal{X}$ and label space $\mathcal{Y}$, and by $\uell_{\infty}(\epsilon) =\{\bm{\delta} \  \vline \  \left\lVert \bm{\delta} \right\rVert _{\infty} \leq \epsilon \}$ a centered $\uell_{\infty}$ ball with perturbation radii $\epsilon >0$. Given any original sample $(\bm{x},y) \in D$, the adversarial attack seeks a perturbation $\bm{\delta}^{\star}$ within $\uell_{\infty}(\epsilon)$ which incurs the largest loss around $\bm{x}$, \emph{i.e.,},
\begin{align}\label{eq:1}
\bm{\delta}^{\star} =\argmax_{\bm{\delta} \in \uell_{\infty}(\epsilon)} L \bigl (f_{\bm{\theta}}(\bm{x}+\bm{\delta}),y \bigr),
\end{align}
where $L(\cdot,\cdot)$ represents the target loss function (e.g. cross entropy) and $f_{\bm{\theta}}:\mathcal{X} \rightarrow \mathcal{Y}$ is a classifier parameterized by $\bm{\theta} \in \bm{\Theta}$. The adversarial sample  $\bm{x} + \bm{\delta}^{\star}$ is used for adversarial training for $f_{\bm{\theta}}$. AT methods search for the robust model $f_{\bm{\theta}^{\star}}$ where
\begin{align}\label{eq:2}
\bm{\theta}^{\star}=\argmin_{\bm{\theta} \in \bm{\Theta}} \mathbb{E}_{(\bm{x},y)\sim D}\left[L(f_{\bm{\theta}}(\bm{x}+\bm{\delta}^{\star}),y)\right],
\end{align}
and $\bm{\theta}^{\star}$ is often approximated iteratively by solving a min-max optimization problem consisted of equation (\ref{eq:1}) and (\ref{eq:2}).

To solve (\ref{eq:1}), single-step methods such as N-FGSM firstly initialize a perturbation $\bm{\delta}^0$ where each of its entries is drawn independently from an uniform distribution $U(-k\epsilon,k\epsilon)$ with an augmentation parameter $k\geq 1$, and then update $\bm{\delta}^0$ using local grad information in $\bm{x}+\bm{\delta}^0$ to obtain a linear estimation $\hat{\bm{\delta}}$ of $\bm{\delta}^{\star}$, \emph{i.e.},
\begin{align}
    \hat{\bm{\delta}} = \bm{\delta}^0+\epsilon \cdot \mathrm{sign} \bigl[ \nabla_{\bm{\delta}^0}L \bigl(f_{\bm{\theta}}\left(\bm{x}+\bm{\delta}^0\right),y \bigr) \bigr]. \notag
\end{align}

Multi-step methods such as PGD-n, however, update $\bm{\delta}^0$ multiple times to get a better approximation of $\bm{\delta}^{\star}$ \cite{zhang2019theoretically}, \emph{i.e.},
\begin{align}
     \hat{\bm{\delta}}\triangleq\bm{\delta}^{t+1}
    = \Pi \Bigl[
    \bm{\delta}^t
    +\alpha \cdot 
    \mathrm{sign} 
    \Bigl(
        \nabla_{\bm{\delta}^t}
        L \bigl(
            f_{\bm{\theta}}\left(
                \bm{x}+\bm{\delta}^t\right)
            ,y \bigr)\Bigr) \Bigr],
     t=0,1,\ldots,n-1, \notag
\end{align}
where $\alpha >0 $ is the step length and $\Pi(\cdot)$ the projection operator onto $\uell_{\infty}(\epsilon)$. For any of the two adversarial attack methods presented here, we would denote by $\bm{\hat{x}} = \bm{x} + \hat{\bm{\delta}}$ the corresponding adversarial sample which obviously depends on the original sample $\bm{x}$ and the initial perturbation $\bm{\delta}^0$.

\section{Our approach}\label{sec:Method}

In this section, we give a detailed description of our framework, a proposed tuned version of LHS aiming to control expenses, and three selection strategies used in experiments.

\subsection{Batch-in-Batch: Duplication for better preparation}

\begin{algorithm}[H]\label{alg: BB}
    \caption{Batch-in-Batch}
    \SetKwInOut{Input}{input}
    \SetKwInOut{Output}{output}
    \Input{original batch $\mathcal{B}_{\mathrm{ori}}$, training model $f_{\bm{\theta}}$, number of duplication $m$,  initial perturbation function $\mathrm{NOISE}$, adversarial attack function $\mathrm{ADV}$ \footnotemark,  sample selection function $\mathrm{SELECT}$. }
    $\mathcal{B}_{\mathrm{bgn}}=\mathrm{CONCATENATE}(\mathcal{B}_{\mathrm{ori}},\mathcal{B}_{\mathrm{ori}},\ldots,\mathcal{B}_{\mathrm{ori}})$ \tcp*{$m$ copies of $\mathcal{B}_{\mathrm{ori}}$.}
    $\bm{\delta}_0=\mathrm{NOISE}(\mathcal{B}_{\mathrm{ori}},m)$ \tcp*{Jointly designed for $m$ copies.}
    $\mathcal{B}_{\mathrm{adv}}=\mathrm{ADV}(\mathcal{B}_{\mathrm{bgn}}+\bm{\delta}_0,f_{\bm{\theta}})$\;
    $\mathcal{B}_{\mathrm{adv}}=\mathrm{CONCATENATE}(\mathcal{B}_{\mathrm{adv}},\mathcal{B}_{\mathrm{ori}})$ \tcp*{An optional step}
    $\mathcal{B}_{\mathrm{f}}=\mathrm{SELECT}(\mathcal{B}_{\mathrm{adv}},f_{\bm{\theta}})$ \tcp*{$f_{\bm{\theta}}$ is used to predict logits.}
    \Output{Final training batch $\mathcal{B}_{\mathrm{f}}$.}
\end{algorithm}\footnotetext{Notice that $\mathrm{ADV}$ should exclude initial perturbation, which has been done by $\mathrm{NOISE}$.}

Algorithm \ref{alg: BB} presents the components and procedure of proposed Batch-in-Batch (BB) framework. More specifically, given a training batch $\mathcal{B}_{\mathrm{ori}}=\{\bm{x}_{1}, \bm{x}_2, \cdots, \bm{x}_{n}\}$ of size $n$, it is firstly duplicated $m$ times to form a larger batch set $\mathcal{B}_{\mathrm{bgn}}=\{\mathcal{B}_{\mathrm{ori}},\mathcal{B}_{\mathrm{ori}},\ldots,\mathcal{B}_{\mathrm{ori}}\}$. Then the initial perturbation function NOISE and the adversarial attack function ADV, are subsequently applied to get the initial perturbation $\bm{\delta}_0$ and the adversarial batch $\mathcal{B}_{\mathrm{adv}}$. Finally, the SELECT strategy would pick adversarial samples in $\mathcal{B}_{\mathrm{adv}}$ based on the predictions of the current training model $f_{\bm{\theta}}$ on $\mathcal{B}_{\mathrm{adv}}$. Paired with a more delicate method to construct initial perturbation (section \ref{subsec:LHS}) and various selection strategies (section \ref{sec:sample selection}), the framework offers adversarial samples with improved diversity, along with a filtering procedure that eliminate 
unnecessary or harmful training samples. Conventional AT methods can be described with specific settings of ADV, $m$, NOISE and SELECT.

\subsection{A tuned LHS for Batch-in-Batch}\label{subsec:LHS}

As multiple copies of the original batch are processed simultaneously, it is possible to design a series of initial perturbations for each of these copies such that they disperse more evenly within the vicinity of original samples. 
To this end, we modify Latin Hypercube Sampling \cite{LHS1979}, a method that was previously challenging to implement in adversarial training, to jointly provide diverse initial values while controlling its computational expenses in high-dimensional feature spaces.

\vspace{-0.4cm}
\begin{algorithm}
    \caption{A tuned Latin hypercube sampling (tLHS) for Batch in Batch}
    \label{alg:LHS}
    \SetKwInOut{Input}{input}
    \SetKwInOut{Output}{output}
    \Input{number of duplication $m$, original batch $\mathcal{B}_{\mathrm{ori}}$, attack radius $\epsilon>0$ ($k\epsilon$ for N-FGSM).}
    $\bm{\gamma}=(0,1/m,2/m,\ldots,(m-1)/m)^{\top}$\;
    $\bm{\delta}=(\delta_{i,j})_{m \times d}$\;
    \For{$j=1,\ldots,d$}{
        $\bm{\delta_{\cdot,j}}=\ $random shuffled $\bm{\gamma}$
        \tcp*[l]{Fill columns with permutations of $\bm{\gamma}$}}
    $\bm{\tau}=(\tau_{i,j})_{m \times d},\, \tau_{i,j} \overset{\text{i.i.d.}}{\sim} U(0,1/m)$\;
    $\bm{\delta}_{\text{LHS}}=2\epsilon(\bm{\delta}+\bm{\tau})-\epsilon$ \tcp*{Perturb and cast entries on $(-\epsilon,\epsilon)$}
    $\bm{\delta}_\mathrm{f}=\mathrm{column\_bind}(\bm{\delta}_{\text{LHS}},\bm{\delta}_{\text{LHS}},\ldots,\bm{\delta}_{\text{LHS}})$ \tcp*{$n$ identical copies of $\bm{\delta}_{\text{LHS}}$}
    Reshape $\bm{\delta}_\mathrm{f}$ to the same size as $\mathcal{B}_{\mathrm{bgn}}$\;
    \Output{Initial perturbation $\bm{\delta}_{\text{f}}$ for the larger batch $\mathcal{B}_{\mathrm{bgn}}$.}
\end{algorithm}
\vspace{-0.4cm}

Given the inputs, Algorithm \ref{alg:LHS} would, at the beginning of each training epoch, firstly yield a matrix $\bm{\delta}$ where each column is a permutation of $\bm{\gamma}$. Then an initial perturbation $\bm{\delta}_{\text{LHS}}$ is cast to the desired range via a linear transformation operated on the sum of $\bm{\delta}$ and random matrix $\bm{\tau}$. It is noticed that $\bm{\delta}_{\text{LHS}}$ has $m$ rows of dimension $d$ and they can be regarded as initial perturbations for $m$ duplications of a certain original sample in $\mathcal{B}_{\mathrm{ori}}$.

To control computational costs of Algorithm \ref{alg:LHS} and to account for each of the $n$ samples in $\mathcal{B}_{\mathrm{ori}}$, we set $\bm{\delta}_{\text{f}}$ as a concatenation of $n$ copies of $\bm{\delta}_{\text{LHS}}$, and reshape it to the same size as $\mathcal{B}_{\mathrm{bgn}}$, therefore obtaining the final initial perturbations $\bm{\delta}_{\text{f}}$ for the larger batch $\mathcal{B}_{\mathrm{bgn}}$. It is worth mentioning that our algorithm will not trigger the same problem as R+FGSM does, where all entries of samples are initialized and projected to the very limited set $\{-1,1\}$. A significant distinction between R+FGSM and the reusing of one set $\bm{\delta}_{\text{LHS}}$ is that the recurrence of $1$ and $-1$ persists throughout all epochs of R+FGSM training, while we update the initial perturbation with tLHS to redistribute all initial positions without repetition to previous epoch, thereby avoiding repetitive entries are used.

Compare to naive random initialization under uniform distribution, initial perturbation yield by Algorithm \ref{alg:LHS} would make the adversarial attack richer in gradient information around original samples, which further benefits the sample selection by providing adversarial samples with better diversity.

\subsection{Sample selection strategies}\label{sec:sample selection}

Since initial perturbation and adversarial attack is performed on $\mathcal{B}_{\mathrm{bgn}}$ which is consisted of $m$ copies of $\mathcal{B}_{\mathrm{ori}}$, the resulting adversarial batch $\mathcal{B}_{\mathrm{adv}}$ would also contains $m$ elements and we denote it by 

\begin{equation*}\label{eq:B_{adv}}
    \mathcal{B}_{\mathrm{adv}} = \left\{ \mathcal{B}_{\mathrm{adv}}^{(1)}, \cdots, \mathcal{B}_{\mathrm{adv}}^{(m)}\right\} =\left\{\begin{array}{
    c c c}
    \centering
     \bm{x}_1^{(1)}     & \ldots & \bm{x}_1^{(m)}  \\
     \bm{x}_2^{(1)}     & \ldots & \bm{x}_2^{(m)}  \\
     \vdots             & \ddots & \vdots  \\
     \bm{x}_{n}^{(1)} & \ldots & \bm{x}_{n}^{(m)} 
\end{array}\right\}=\left\{
  \begin{array}{c}
          \bm{r}_{1} \\
          \bm{r}_{2} \\
          \vdots \\
          \bm{r}_{n}
 \end{array}
 \right\},
\end{equation*}
where $\bm{x}_i^{(j)}$ is the $j$-th adversarial sample of original sample $\bm{x}_i \in \mathcal{B}_{\mathrm{ori}}$, and each row $\bm{r}_{i}=\{\bm{x}_i^{(1)},\ldots,\bm{x}_i^{(m)} \}$ of $\mathcal{B}_{\mathrm{adv}}$ includes all $m$ adversarial samples related to $\bm{x}_{i}$.

To further get the final training sample $\mathcal{B}_{\mathrm{f}}$ from $\mathcal{B}_{\mathrm{adv}}$, we begin from one selection strategy (referred to as CP afterwards) originated from Kim et al. \cite{kim2021understanding}, and then propose two other strategies from the perspectives of \emph{greedy search} and \emph{datum-wise early stopping}. In CP strategy, given the training model $f_{\bm{\theta}}$ and $\bm{r}_i$, it picks out the misclassified adversarial sample (if exists) with minimum $\uell_{\infty}$ norm to original sample, \emph{i.e.,}
\begin{align} \label{eq:CP}
    \mathrm{SELECT}_{\mathrm{CP}}
    \left(\bm{r}_i,f_{\bm{\theta}}\right)=
    \left\{
    \begin{aligned}
        \argmin_{j \in M_i} \left\lVert \bm{x}_i^{(j)}-\bm{x}_i \right\rVert_{\infty}, \text{if} \,\, M_i\neq \emptyset,\\
        \argmax_{1\leq j \leq m} \left\lVert \bm{x}_i^{(j)}-\bm{x}_i \right\rVert_{\infty}, \text{if} \,\, M_i=\emptyset,
    \end{aligned}
    \right.
\end{align}
where $M_i=\{j|f_{\bm{\theta}}(\bm{x}_i^{(j)}) \neq y_i\}$ denotes indexes of all misclassified adversarial samples in $\bm{r}_{i}$.  And $\mathcal{B}_{\mathrm{f}}$ is obtained by applying CP on each $\bm{r}_{i}$ of $\mathcal{B}_{\mathrm{adv}}$. The main difference between original CP strategy and ours implementation is that adversarial samples $\{\bm{x}_i^{(j)}, j=1,\ldots,m\}$ are not collinearly aligned and the sample pool to be selected from contains gradient information from multiple positions.

Motivated by CP selection and early stopping, we propose a new Greedy Selection (GS) strategy that chooses only misclassified adversarial samples from $\mathcal{B}_{\mathrm{adv}}$ to avoid potentially unnecessary or harmful training on a per-datum level, and the final training samples $\mathcal{B}_{\mathrm{f}}$ is obtained by
\begin{align}\label{eq:GS}
\mathcal{B_{\mathrm{f}}}=\mathrm{SELECT}_{\mathrm{GS}}\left(\mathcal{B}_{\mathrm{adv}},f_{\bm{\theta}}\right)=\left\{\bm{x}|\bm{x} \in \mathcal{B}_{\mathrm{adv}}, \bm{x} \ \mathrm{is\ misclassified\ by\ } f_{\bm{\theta}}\right\}. 
\end{align}

One can notice that the major difference between CP and GS is that for each original sample $\bm{x}_{i}, i=1, 2, \cdots, n$, CP would  select only one adversarial sample from its $m$ related adversarial samples $\{\bm{x}_{i}^{(j)}, j=1, 2, \cdots, m\}$ while GS could select zero or multiple examples from it. The intuition behind GS is that if all adversarial attack attempts failed on a certain original sample, then additional training for those adversarial samples is unnecessary. In Section \ref{sec:Experiments}, it is shown that this greedy strategy greatly improves model robustness in many settings. We also explain the mechanism behind it with experiments from multiple perspectives.

The third selection strategy, termed as Balanced Greedy (BG), modifies GS by replacing $\mathcal{B_{\mathrm{adv}}}$ in equation (\ref{eq:GS}) with a larger set

\begin{equation}\label{eq:B_{S}}
    \mathcal{B}_{\mathrm{S}} = \left\{\mathcal{B}_{\mathrm{ori}},\mathcal{B}_{\mathrm{adv}}\right\}. 
\end{equation}
Hence, BG strategy would enable sample selection from not only clean original samples but also perturbed adversarial samples, and the number $m$ of duplication could play a role in controlling the trade-off between clean and adversarial accuracy. Specifically, when $\mathcal{B}_{\mathrm{f}}$ is empty in the second or third strategy, training for $\mathcal{B}_{\mathrm{ori}}$ will be skipped (which rarely occurs).

\section{Experiments}\label{sec:Experiments}

In this section, we test the performance of our Batch-in-Batch framework on three datasets (CIFAR-10, CIFAR-100 \cite{krizhevsky2009learning} and SVHN \cite{netzer2011reading}) with different initializing perturbation methods (Algorithm \ref{alg:LHS} and simple uniform sampling), adversarial attack methods (N-FGSM and PGD-10), sample selection strategies (CP, GS and BG), and with two model architectures (PreactResNet18 \cite{he2016identity} and WideResNet28-10 \cite{zagoruyko2016wide}). We utilize Algorithm \ref{alg:LHS} as the NOISE function for initial perturbation within our framework by default unless otherwise specified, while retaining the original naive uniform random initialization as NOISE for N-FGSM and PGD-10. The model training settings closely resemble those used by \cite{kim2021understanding}, with the exception of reducing the training epochs from 200 to 75 due to computational constraints. Specifically, we choose \textit{stochastic gradient descent} (SGD) \cite{ruder2016overview} with a learning rate of 0.01, a momentum of 0.9, and a weight decay of 5e-4. Additionally, the learning rate is decayed with a factor of 0.1 at 28 and 56 epochs.

For single-step method N-FGSM, the radii augmentation parameter $k$ is set to be 2, as the default setting in the original paper \cite{jorge2022make}. And for multi-step method PGD-10, the step length is $\alpha=\epsilon/4$ where $\epsilon$ is the attack radius. The adversarial accuracy is estimated by performing PGD-50 on the test set of each dataset following the approach used in previous studies \cite{kim2021understanding, jorge2022make, Gu2022SegPGDAE}. We report the best adversarial accuracy on the test set along with its corresponding clean accuracy. Part of the results can be found in Table \ref{table:table 1} and full results in appendix \ref{ap:1}. Our code is available at \url{https://github.com/Yinting-Wu/Batch-in-Batch}.

\subsection{Compared to plain N-FGSM and PGD-10}\label{subsec:ex1}

\begin{table}[h!]
\centering
\caption{Table \ref{table:table 1} presents the clean and adversarial accuracy of two baseline models and six models trained within the BB framework, equipped with tLHS as initial perturbation for adversarial attack, and with one of the three sample selection strategies. The two results in each cell are respectively the mean clean accuracy (top) and mean adversarial accuracy (bottom) with their standard deviations computed on three random seed. The backbone network of classifier $f_{\theta}$ is WRN-28}
\label{table:table 1}
\begin{tabular}{c|m{4.2em}<{\centering}|m{4.2em}<{\centering}|m{4.2em}<{\centering}|m{4.2em}<{\centering}|m{4.2em}<{\centering}|m{4.2em}<{\centering}}
\toprule
Dataset & \multicolumn{2}{c}{CIFAR-10} & \multicolumn{2}{c}{SVHN} & \multicolumn{2}{c}{CIFAR-100} \\
$\epsilon$ & \multicolumn{1}{c}{8/255} & \multicolumn{1}{c}{16/255} & \multicolumn{1}{c}{4/255} & \multicolumn{1}{c}{8/255} & \multicolumn{1}{c}{8/255} & \multicolumn{1}{c}{16/255} \\
\hline\hline
\multirow{2}{8em}{N-FGSM-plain} 
    & $77.15_{\pm2.02}$ & $59.28_{\pm0.65}$ 
    & $95.39_{\pm0.82}$ & $91.48_{\pm0.13}$ 
    & $48.75_{\pm0.65}$ & $30.08_{\pm0.42}$ \\
 & $44.82_{\pm0.19}$ & $\bm{26.17}_{\pm0.76}$ 
 & $71.17_{\pm0.67}$ & $42.03_{\pm0.79}$ 
 & $23.75_{\pm0.45}$ & $12.81_{\pm0.87}$ \\
\hline
\multirow{2}{8em}{BB(CP)-N-FGSM} 
& $79.22_{\pm0.27}$ & $58.83_{\pm0.80}$ 
& $95.31_{\pm0.53}$ & $89.84_{\pm0.43}$ 
& $49.02_{\pm0.66}$ & $34.57_{\pm0.74}$ \\
 & $43.91_{\pm0.72}$ & $24.45_{\pm0.91}$ 
 & $70.21_{\pm0.93}$ & $42.56_{\pm0.19}$ & 
 $\bm{24.32}_{\pm0.72}$ & $\bm{13.48}_{\pm0.48}$ \\
\hline
\multirow{2}{8em}{BB(GS)-N-FGSM} 
& $76.09_{\pm0.05}$ & $62.27_{\pm0.87}$ 
& $93.95_{\pm0.10}$ & $90.16_{\pm0.96}$ 
& $51.66_{\pm0.50}$ & $33.79_{\pm0.80}$ \\
 & $\bm{47.11}_{\pm0.21}$ & $26.02_{\pm0.64}$ 
 & $78.61_{\pm0.47}$ & $\bm{55.37}_{\pm0.07}$ 
 & $23.14_{\pm0.81}$ & $12.70_{\pm0.30}$ \\
\hline
\multirow{2}{8em}{BB(BG)-N-FGSM} 
& $79.14_{\pm0.38}$ & $57.19_{\pm0.07}$ 
& $94.53_{\pm0.85}$ & $94.37_{\pm0.55}$ 
& $53.71_{\pm0.78}$ & $29.88_{\pm0.03}$ \\
 & $46.72_{\pm0.13}$ & $23.28_{\pm0.06}$ 
 & $\bm{79.10}_{\pm0.44}$ 
 & $51.48_{\pm0.90}$ & $21.39_{\pm0.86}$ & $8.89_{\pm1.10}$ \\
\midrule
\multirow{2}{8em}{PGD-10-plain} 
& $77.89_{\pm0.97}$ & $58.67_{\pm0.18}$ 
& $95.62_{\pm0.50}$ & $92.58_{\pm0.30}$ 
& $51.48_{\pm0.55}$ & $32.89_{\pm0.38}$ \\
 & $46.33_{\pm0.50}$ & $27.42_{\pm0.35}$ 
 & $73.44_{\pm0.20}$ & $51.64_{\pm0.22}$ 
 & $25.23_{\pm0.69}$ & $13.75_{\pm0.12}$ \\
\hline
\multirow{2}{8em}{BB(CP)-PGD-10} 
& $78.61_{\pm1.04}$ & $58.89_{\pm0.31}$ 
& $96.09_{\pm0.17}$ & $91.72_{\pm0.62}$ 
& $50.47_{\pm0.44}$ & $32.66_{\pm0.57}$ \\
 & $46.39_{\pm1.16}$ & $29.10_{\pm0.29}$ 
 & $73.83_{\pm0.81}$ & $50.78_{\pm0.23}$ 
 & $\bm{25.76}_{\pm0.25}$  & $\bm{13.98}_{\pm0.71}$ \\
\hline
\multirow{2}{8em}{BB(GS)-PGD-10} 
& $76.56_{\pm0.82}$ & $61.33_{\pm0.51}$ 
& $93.59_{\pm0.35}$ & $91.33_{\pm0.05}$ 
& $50.31_{\pm0.37}$ & $31.80_{\pm0.91}$ \\
 & $\bm{49.61}_{\pm0.96}$ & $\bm{30.76}_{\pm0.54}$ 
 & $\bm{83.20}_{\pm0.28}$ & $59.38_{\pm0.15}$ 
 & $23.44_{\pm0.46}$ & $13.52_{\pm0.06}$ \\
\hline
\multirow{2}{8em}{BB(BG)-PGD-10} 
& $77.54_{\pm0.67}$ & $63.87_{\pm0.89}$ 
& $94.37_{\pm0.56}$ & $91.61_{\pm0.39}$ 
& $52.03_{\pm0.89}$ & $38.36_{\pm0.83}$ \\
 & $48.34_{\pm0.42}$ & $29.30_{\pm0.19}$ 
 & $81.25_{\pm0.21}$ & $\bm{60.08}_{\pm0.31}$ 
 & $22.27_{\pm0.86}$ & $12.27_{\pm0.48}$ \\
\bottomrule
\end{tabular}
\end{table}

We test the performance of our Batch-in-Batch framework on three datasets with proposed initial perturbation (\emph{i.e.,} Algorithm \ref{alg:LHS}) and sample selection strategies discussed in Section \ref{sec:sample selection}. The two 
adversarial attack methods (\emph{i.e.,} ADV in Algorithm \ref{alg: BB}) to be considered are the N-FGSM and PGD-10. Table \ref{table:table 1} gives both the clean and adversarial accuracy of six models trained using the BB framework with varying attack radii $\epsilon$, where N-FGSM-plain and PGD-10-plain are the two baseline AT methods with uniform initial perturbation and without sample selection. The backbone network of classifier $f_{\bm{\theta}}$ is WideResNet28-10 (abbreviated by WRN-28 afterwards) (experiments with PreActResNet18 network, abbreviated by PRN-18 afterwards, are given in Table \ref{table:SVHN}, \ref{table:CIFAR-10}, and \ref{table:CIFAR-100}).

\textbf{Performance of the checkpoint selection strategy (CP).} When comparing with baselines, models trained based on CP strategy consistently provide a slight improvement in adversarial accuracy while maintaining comparable clean accuracy in most settings. Besides, CP achieves the highest adversarial accuracy on the CIFAR-100 dataset, which has fewer samples for each class compared to CIFAR-10 and SVHN. We argue that this can be explained by two reasons. The first one, as described in Section \ref{sec:sample selection}, is that CP strategy would not ignore the selection (or information) of adversarial samples for any of the original samples, and this becomes crucial especially when the original samples within each class are scarce. The second reason is that the size of the final training samples $\mathcal{B_{\mathrm{ori}}}$, obtained with CP, remains constant (\emph{i.e.,} $m$) during the training process, whereas this value is dynamic and tends to be larger with the other two strategies, which is believed to potentially have a negative impact on training \cite{keskar2016large, lin2018don}.

\textbf{Performance of the greedy selection strategy (GS).} When comparing the adversarial accuracy of the BB(GS)-PGD-10 with that of PGD-10-plain, it is observed that the former significantly improves the adversarial accuracy on both the CIFAR-10 and SVHN dataset, with only little price to pay on the clean accuracy. Similarly, the BB(GS)-N-FGSM also outperforms the N-FGSM-plain in almost all $\epsilon$ scenarios concerning the CIFAR-10 or SVHN dataset. And the improvement in adversarial accuracy can reach up to 13.34\% (\emph{i.e.,} 55.37\% vs 42.03\% on SVHN and $\epsilon=8/255$), while maintaining the clean accuracy almost unchanged to baseline model. On the CIFAR-100 dataset, models trained with the GS strategy fail to increase adversarial accuracy. We argue that this is because the GS strategy does not accurately filter out unnecessary training samples, occasionally eliminating useful ones. This may prevent models from learning sufficient features, especially on CIFAR-100, where each class has relatively fewer samples, making the loss of valuable training data more detrimental compared to the other two datasets.

\textbf{Performance of the balanced greedy strategy (BG).} Same as the original envision, the BG strategy do play the role of balancing clean and adversarial accuracy compared to the GS strategy. More precisely, BG enables models to yield higher clean accuracy but slightly lower adversarial accuracy than their GS-based competitors in almost all settings considered in Table \ref{table:table 1}. The reason behind it is straightforward since BG strategy allows the selection of clean original samples from $\mathcal{B}_{\mathrm{ori}}$ to be included in the final training samples $\mathcal{B}_{\mathrm{f}}$ for the model under consideration, enabling the model to pay more attention to the clean examples during the training. In addition, similar to the performance of the GS strategy, the models trained with BG also achieve more significant improvement of the results on datasets with more per-class samples when competing with baseline models. 

\subsection{Analysis of the efficiency of sample selection strategies}\label{subsec:ex2}

In this section, we conduct multiple experiments from various perspectives to closely examine how the greedy selection strategy, which is based on datum-wise early stopping, improves adversarial accuracy.

\subsubsection{Impacts on loss landscapes}

\begin{figure}[hbt!]
\includegraphics[width=30em]{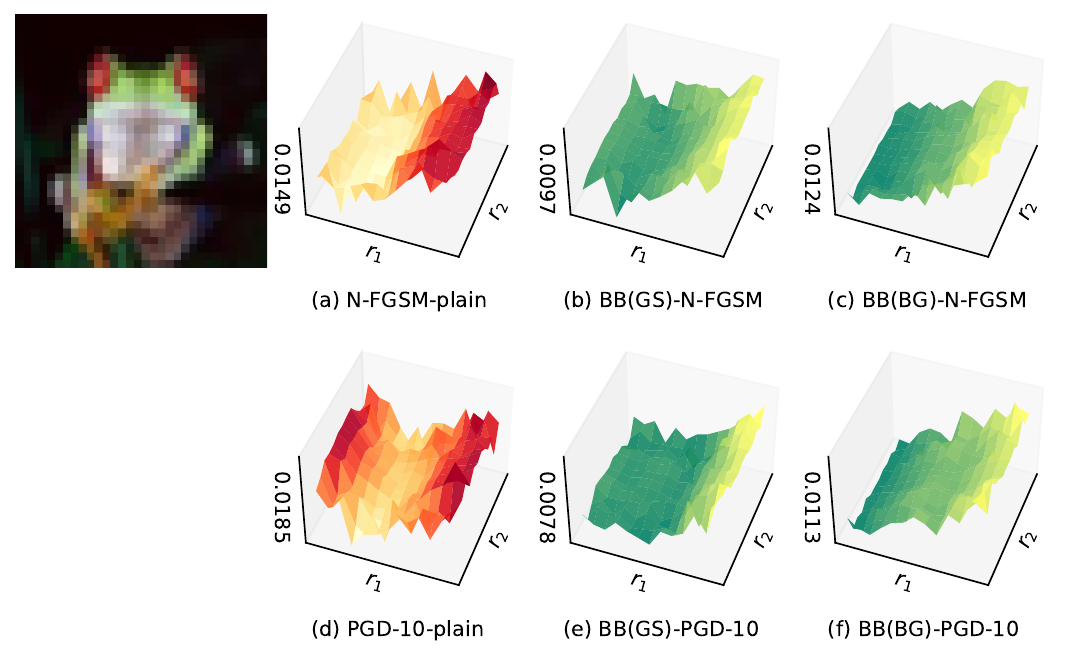}
\centering
\caption{Loss landscapes of six models considered in Table \ref{table:table 1} where the first row corresponds to losses of N-FGSM based models while the second losses of PGD-10 based models. Followed the methodology described in \cite{engstrom2018evaluating}, local losses are calculated as $L(f_{\bm{\theta}}(\bm{x}+t_{1}\bm{r}_1+t_{2}\bm{r}_2),y)$ where $(\bm{x}, y)$ is an original CIFAR-10 image (\emph{i.e.,} frog) with its label; $\bm{r}_1=\text{sign}[\nabla_{\bm{x}}L(f_{\bm{\theta}}(\bm{x}),y)]$ is the gradient direction and $\bm{r}_2 \sim \mathrm{Rademacher}(0.5)$ is a random direction;  $t_1$ and $t_2$ are evenly distributed scalars on [-0.1, 0.1].
Numbers on the z-axis of each subplot represent standard deviation of losses}
\vspace{-0.6cm}
\label{fig:loss_surface_both}
\end{figure}

We start by plotting local losses around an original point for models trained under different settings since local linearity of the loss function makes it easier for adversarial attack to find the local loss maximum via gradient approximation. Better linearity forces the training model to confront with more aggressive adversarial samples, leading to improved robustness \cite{andriushchenko2020understanding}. Figure \ref{fig:loss_surface_both} shows loss landscapes of six models considered in Table \ref{table:table 1} where the first (\emph{resp} second) row of landscapes represents N-FGSM (\emph{resp} PGD-10) based models. Landscapes of more models and can be found in appendix \ref{secA1}. It is found that loss surface of N-FGSM based (\emph{resp} PGD-10) models trained within BB framework with either GS or BG selection strategy achieves smoother and more linear trend on adversarial direction $\bm{r}_{1}$ than that of N-FGSM-plain (\emph{resp} PGD-10-plain), and has much smaller standard deviation. This phenomenon could be attributed to avoiding training on correctly classified samples, which is the fundamental concept behind GS and BG. Moreover, it not only implies that models trained with GS or BG would not fluctuate drastically when inputs are slightly perturbed but also demonstrates the effectiveness of these two strategies proposed in the BB framework in enhancing the robustness of trained models.

\subsubsection{Consistency of attack success rate between single and multi-step attack}

\begin{figure}[hbt!]
\includegraphics[width=\linewidth]{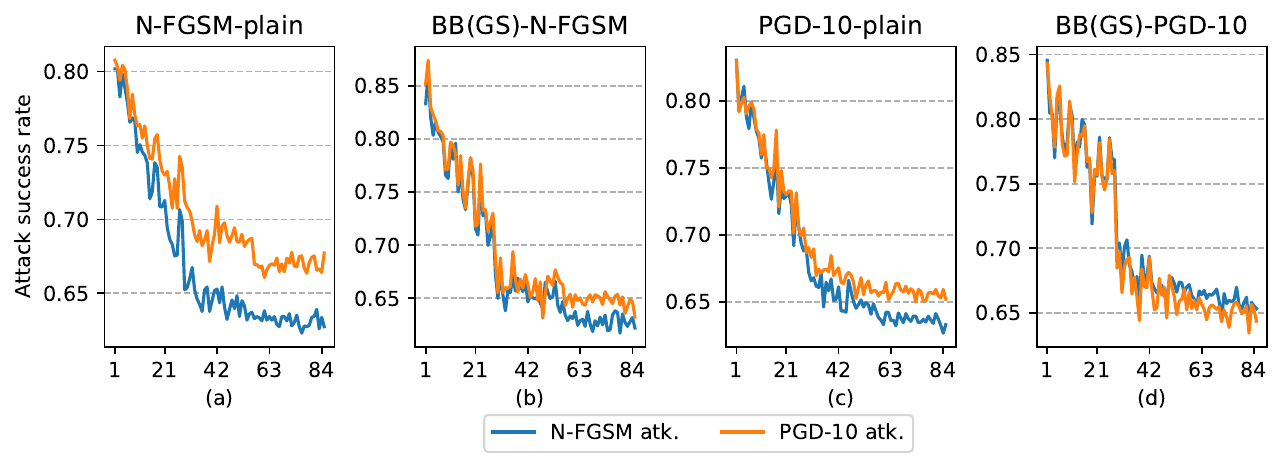}
\centering
\caption{The success rates of both N-FGSM attack (blue line) and PGD-10 attack (orange line) against two baseline models (N-FGSM-plain and PGD-10-plain) and two models (BB(GS)-N-FGSM and BB(GS)-PGD-10) trained in BB framework. The success rate of two attacks for each model are calculated on the training set of CIFAR-10 dataset during the training process}
\label{fig:success_rate}
\end{figure}

Multi-step attacks generally yield a higher attack success rate than single-step attacks because they achieve a better approximation to the local maximum of the loss function through iterative gradient computations. If the loss function of a certain model is ideally linear enough, single and multi-step attacks would, however, have a similar success rate since the former can effectively approximate the local maximum in a single round of gradient computation. To compare the smoothness of trained models from this perspective, we monitored the success rates of both single (N-FGSM) and multi-step (PGD-10) attacks on each of the four models considered above during the training process on the CIFAR-10 dataset (Figure \ref{fig:success_rate}).

For both the N-FGSM-plain and PGD-10-plain models, the success rate of a multi-step attack against them becomes much larger than that of a single-step attack as the training progresses, illustrating that the two models are more vulnerable to multi-step attacks. This indicates the fluctuation of the loss surface of the models which makes the single-step attacks fail to effectively find linear approximation of local maxima as multi-step attacks do. However, a similar trend does not appear for either the BB(GD)-N-FGSM or BB(GS)-PGD-10 model, indicating that models trained within our framework have more linear loss landscapes, which facilitate N-FGSM in achieving a success rate close to that of PGD-10. Moreover, the slightly higher success rate of the N-FGSM attack against the BB(GS)-PGD-10 model in the final stage of training can be explained by the attack radius augmentation procedure in N-GFSM, which leads to a wider perturbation range compared to other attack methods.

\vspace{-0.8cm}

\subsubsection{Comparison of models' confidence on predictions}

\begin{figure}[hbt!]
\vspace{-0.5cm}
\includegraphics[width=\linewidth]{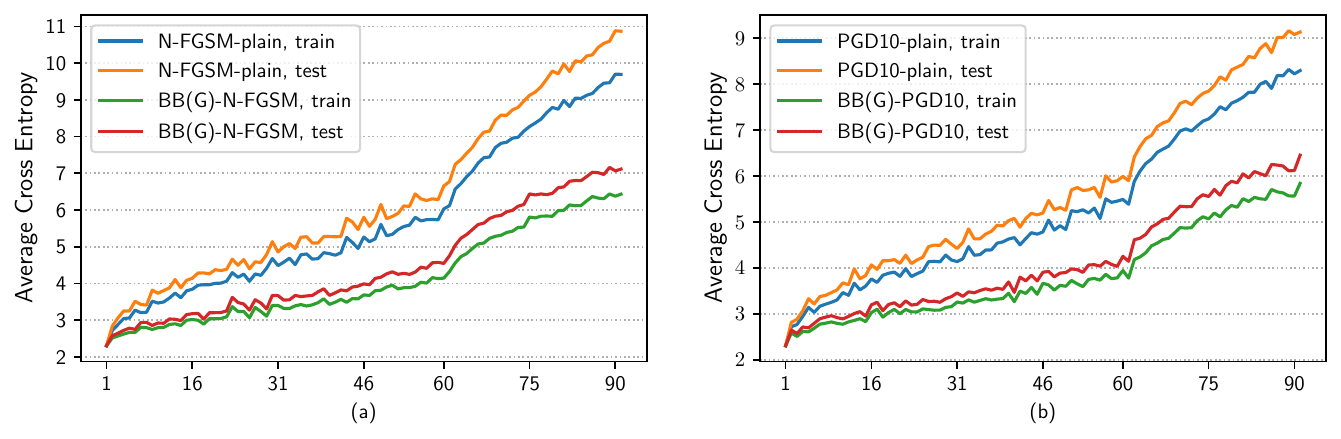}
\centering
\caption{The left (\emph{resp} right) figure presents the mean cross entropy values of plain N-FGSM (\emph{resp} PGD-10) model and its BB version during training process, where the mean values are computed on both the train and test set of CIFAR-10 dataset with attack radius $\epsilon=8/255$}
\vspace{-0.4cm}
\label{fig:smoth}

\end{figure}

Overconfident or assertive models would have poorer generalization \cite{hein2019relu, Grabinski2022RobustMA, fu2020label}. Enlightened by the works \cite{chen2021robust, szegedy2016rethinking, shafahi2019label, goibert2019adversarial, stutz2020confidence} about label smoothing in mitigating model's overconfidence, we considered the cross entropy between predicted and uniform distribution over classes as a measurement of model's confidence over its predictions. Figure \ref{fig:smoth} presents respectively the mean value of cross entropy of two baseline models and their BB version. It is obvious that models trained within BB framework are much less assertive since they have smaller cross entropy value on both train and test set than their baseline competitors.  

\subsection{Ablation study for initial perturbation}\label{subsec:ex5}

In this section, we provide ablation study for Algorithm \ref{alg:LHS}, the tLHS for initial perturbation. 

\subsubsection{Boosts on accuracy}

We perform ablation experiment for Algorithm $\ref{alg:LHS}$, the tLHS for initial perturbation, to see how it contributes to final results of models trained with the BB framework. Table \ref{table:LHS_ablation} compares the results of two models mentioned in Table \ref{table:table 1} when they are trained with or without tLHS. The clean and adversarial accuracy are computed on the CIFAR-10 dataset, with classifier PRN-18 and GS strategy (refer to the Table \ref{table:CIFAR-10} for complete results). It is noted that the tLHS brings a minor boost in adversarial and clean accuracy, with only a very small increase in training time. It shows that the proposed algorithm can be employed as a beneficial plug-in for our framework.

\begin{table}[h!]
\centering
\caption{Ablation experiments for the tLHS, where the attack radius $\epsilon=16/255$}
\label{table:LHS_ablation}
\begin{tabular}{
    l|m{4em}<{\centering}|m{5em}<{\centering}|m{4em}<{\centering}|m{4em}<{\centering}|m{5em}<{\centering}|m{4em}<{\centering}}
\toprule
& \multicolumn{3}{c|}{BB(GS)-N-FGSM} & \multicolumn{3}{c}{BB(GS)-PGD-10} \\
\cmidrule{2-7}
& \multicolumn{1}{c}{\makecell{Clean \\ accuracy}} & \multicolumn{1}{c}{\makecell{Adversarial \\ accuracy}} & \multicolumn{1}{c|}{\makecell{Step \\ time}} & \multicolumn{1}{c}{\makecell{Clean \\ accuracy}} & \multicolumn{1}{c}{\makecell{Adversarial \\ accuracy}} & \multicolumn{1}{c}{\makecell{Step \\ time}}  \\
\hline
no tLHS & 60.25 & 28.91 & 0.1524 & 60.45 & 32.52 & 0.6431 \\
\hline
with tLHS & 60.70 & 29.14 & 0.1602 & 62.32 & 33.36 & 0.6513 \\
\bottomrule
\end{tabular}
\end{table}

\subsubsection{Filling the feature space better}

\begin{figure}[h!]
\includegraphics[width=\linewidth]{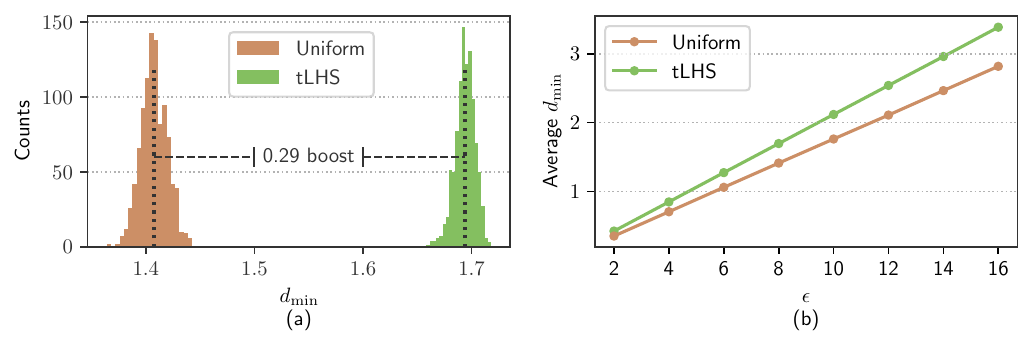}
\centering
\vspace{-0.4 cm}
\caption{Comparison of $d_{\mathrm{min}}$, the minimum $l_{2}$-distance among 3 samples, generated respectively by the tLHS and simple uniform sampling. The histogram (left) of $d_{\mathrm{min}}$ is drawn over 1,000 random seeds, and the line plot (right) shows how $d_{\mathrm{min}}$ of the two methods change with different $\epsilon$, and $d_{\mathrm{min}}$ is estimated on 20 random seeds}
\label{fig:LHS}
\vspace{-0.5cm}
\end{figure}

To examine the effectiveness of Algorithm \ref{alg:LHS}, the tLHS in producing more scattered initial perturbations than simple uniform sampling, we start by generating, for each of these two initialization methods, three samples on the interval $(-\epsilon,\epsilon)^{d}$ where $d$ is the dimension of the feature space. Then $d_{\mathrm{min}}$, the minimum $l_{2}$-distance among samples, are computed respectively for each method, as in work \cite{johnson1990minimax,joseph2008orthogonal}. In Figure \ref{fig:LHS}, the left plot shows the histogram of $d_{\mathrm{min}}$, calculated with 1,000 random seeds and with attack radius $\epsilon=8/255$, and the right line plot presents how average $d_{\mathrm{min}}$ of two methods changes with respect to different values of $\epsilon$. One can observe that our method would produce more diverse initial perturbation for adversarial attacks than uniform sampling does, and that this deviation become more obvious when $\epsilon$ increases.   

\subsubsection{Increasing sample diversity}

\begin{figure}[h!]
\includegraphics[width=\linewidth]{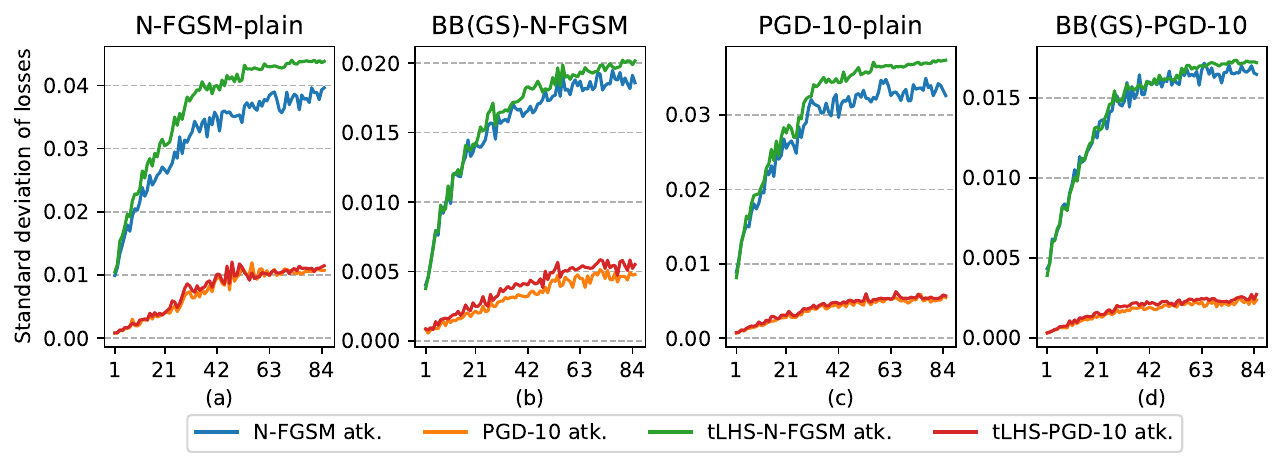}
\vspace{-0.5cm}
\centering
\caption{Monitoring the standard deviation of losses for four models, where each subplot corresponds to the results of a model when it is confronted with four types of attack (colored lines) during its training process. The four types of attack are the combination of two initial perturbations, namely tLHS and uniform sampling, with two attack methods. For example, N-FGSM atk. and tLHS-N-FGSM atk. represent respectively N-FGSM attack method initialized with uniform sampling or tLHS. For each original sample in CIFAR-10 dataset, we generate 4 adversarial samples to calculate the per sample standard deviation. The standard deviation of the loss for a model is then calculated by the mean value of per sample standard deviations on all training samples}
\vspace{-0.6cm}
\label{fig:loss_diversity}
\end{figure}

The losses of adversarial sample, particularly those generated by single-step adversarial training, are highly sensitive to gradient information determined by initial perturbations. When richer gradient information is provided by more diverse initial perturbations, it would be easier for adversarial attacks to efficiently approximate the local maxima of the loss function, resulting in a higher standard deviation of the losses. To verify whether the tLHS is capable of increasing sample's diversity, we will firstly combine two initial perturbations (\emph{i.e.,} tLHS and uniform sampling) with two attack methods (\emph{i.e.,} N-FGSM and PGD-10) to create four types of attack, and then compare the standard deviation of losses when the models considered are confronted with these attacks. The Figure \ref{fig:loss_diversity} presents the corresponding results of four models under consideration, where the colored lines represent respectively four types of attack. It indicates that the attacks initialized with the tLHS achieve a higher standard deviation in adversarial sample losses across the four models, suggesting a higher probability to get better approximation of the local maxima. Notably, models trained with BB(GS) exhibit smaller gaps between two types of single-step attack (\emph{i.e.,} blue and green line). Additionally, the standard deviation of the models trained with BB(GS) is significantly lower than that of the corresponding baseline models. These observations suggest that adversarial samples generated against models trained using our framework are less sensitive to initial perturbations, indicating smoother loss surfaces. These findings align with the conclusions presented in Section \ref{subsec:ex2}.

\subsection{Effects of the hyperparameter $m$}\label{subsec:ex3}

In this section, we presents the effects of $m$ on the adversarial and clean accurcy, and on the training time.
\subsubsection{On two types of accuracy}

\begin{figure}[ht]
\vspace{-0.2 cm}
\includegraphics[width=\linewidth]{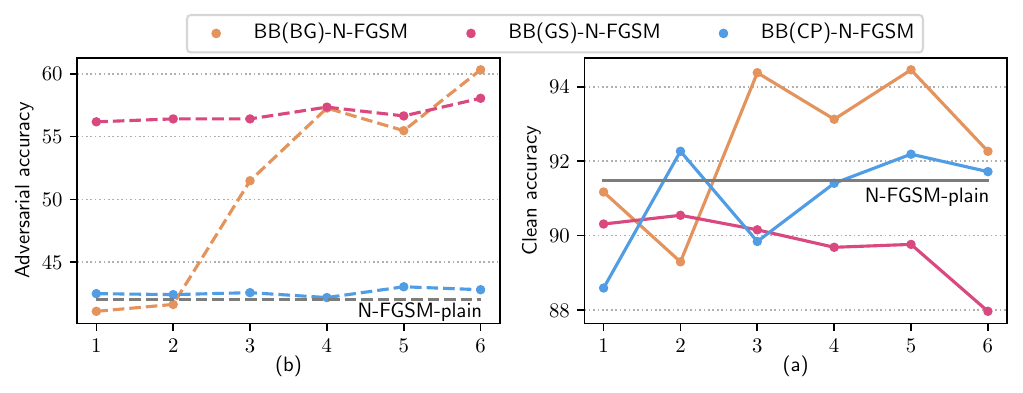}
\centering
\vspace{-0.5 cm}
\caption{The left and right figure gives respectively how adversarial and clean accuracy of three models varies with different values of $m$, where the accuracies are estimated on SVHN dataset with three random seeds, and the attack radius $\epsilon$ equals to $8/255$ with backbone network WRN-28. Although irrespective to $m$, the adversarial and clean accuracy (\emph{i.e.,} the gray line) of the baseline model N-FGSM-plain are also represented in the left and right figure for comparison}
\label{fig:dup ablation}
\vspace{-0.5cm}
\end{figure}

The hyperparameter $m$ decides the number of duplication of original samples in our framework and it is interesting to see how it would affect the performance of models. To this end, we draw in Figure \ref{fig:dup ablation} both adversarial and clean accuracy of four models with different value of $m$. The two types of accuracy are estimated on SVHN dataset with three random seeds, and the attack radius $\epsilon$ equals to $8/255$ with backbone network WRN-28. As $m$ increases, models trained based on any of the three selection strategies eventually outperform the baseline model (\emph{i.e.,} gray dashed line) in adversarial accuracy. Moreover, it is noticed that CP-based model (\emph{i.e.,} blue dashed line) is the least sensitive to $m$ and leads to minor enhancement of the results. This is in accordance with the mechanism of the CP strategy, as it would always select one adversarial example from $\bm{r}_{i}$ (Equation \ref{eq:B_{adv}}), the set containing $m$ adversarial samples related to an original sample $\bm{x}_{i}$, regardless of the size of $m$. In addition, adversarial accuracy of BG-based model (\emph{i.e.,} orange dasehd line) gradually catches up to that of the GS-based model (\emph{i.e.,} pink dashed line) when $m$ increases. The reason behind it is that $\mathcal{B}_{\mathrm{adv}}$ (Equation (\ref{eq:B_{adv}})), the candidate set for GS strategy, would dominate the candidate set $\mathcal{B}_{\mathrm{S}}$ (Equation (\ref{eq:B_{S}})) for BS strategy when $m$ is large.
As for clean accuracy, an increasing $m$ would gradually degrade the performance of GS-based model while enhancing that of BG-based model. This is due to the fact that the BG strategy enables the selection of clean original samples to monitor the training process, thereby preventing the model from training solely on perturbed adversarial samples, as is the case with the the GS strategy.

\subsubsection{On the training time}

\begin{figure}[hbt!]
\vspace{-0.3 cm}
\includegraphics[width=\linewidth]{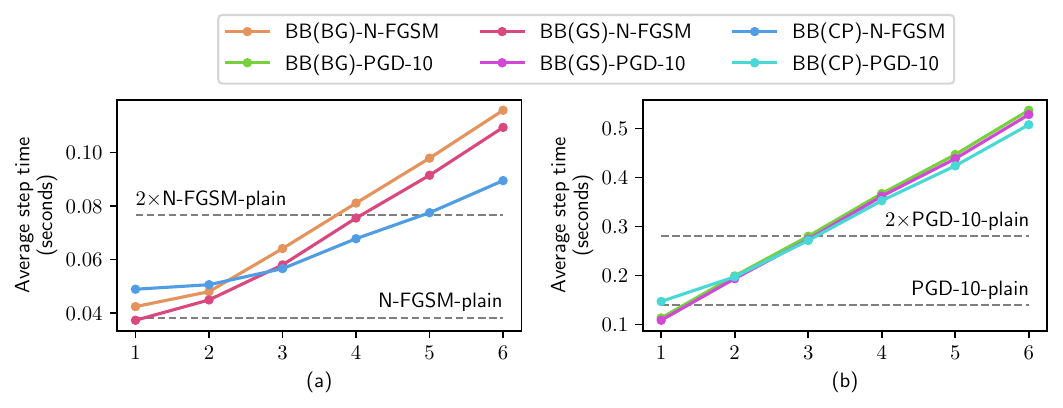}
\centering
\vspace{-0.5 cm}
\caption{The effect of $m$ on the average time per training step for models considered in Table \ref{table:table 1}. The training time for each step is tested on the CIFAR-10 dataset with the PRN-18 network as classifier}
\vspace{-0.4cm}
\label{fig:time ablation}
\end{figure}

To see the computation cost of our framework is cost-effective with different choice of $m$, we recorded, in Figure \ref{fig:time ablation}, the average time for each step in the training of different models considered in Table \ref{table:table 1}. We represent the time cost of baseline models (\emph{i.e.,} N-FGSM-plain and PGD-10-plain) by the gray dashed line for comparison. Although duplicating the original batch $m$ times within our BB framework, it is noted that the computational cost of BB-based model is less than $m$ times that of the baseline counterpart regardless of whether single or multi-step attack method is used. For example, time cost of BB(GS)-N-FGSM model with $m=4$ is approximately twice as expensive as that of N-FGSM-plain model. The explanations for it include that the adversarial samples are generated simultaneously in our framework, and that the size of final training batch $\mathcal{B_{\mathrm{f}}}$ tends to be smaller when the classifier gets better discrimination ability in later training stages.

\section{Conclusions}\label{sec2}

In this work, we aim to improve adversarial training with a combined framework equipped with a  joint initialization and sample selection. Drawing insights from prior research, we hypothesized that some training in adversarial samples might be redundant, and that a joint design of initial perturbation might benefit adversarial training by providing samples with better diversity. Motivated by these two ideas, we propose a novel  training framework that facilitates parallel sample generation for initial perturbation and subsequent sample selection.

Through extensive experimentation, we evaluated the efficacy of our framework and delved into its underlying mechanisms. We observed the impacts of initial perturbation design and sample selection strategies from various perspectives. We find that more considerate initialization improves sample diversity, especially for single-step attack methods like N-FGSM. We also find that when abundant samples are provided for each class, selectively filtering out potentially unnecessary samples during training is able to significantly boost model robustness while keeping clean accuracy almost untouched.

Nevertheless, our framework currently exhibits a notable sensitivity to several configurations such as dataset scarcity, attack radius, and model architecture. We posit that this sensitivity may stem from the somewhat rudimentary nature of our sample filtering mechanism, warranting further refinement. Moving forward, a promising avenue for future research involves exploring more sophisticated schemes for evaluating the utility of training on specific samples. This refinement could lead to a more resilient and adaptable adversarial training framework across various settings and scenarios.

\section{Acknowledgement}

This work is supported by National Natural Science Foundation of China under Grant No. 62107021 and Grant No.61971316, the Fundamental Research Funds for the Central Universities under Grant No. CCNU22JC027 and Knowledge Innovation of Wuhan Science and Technology Project of Wuhan Science and Technology Bureau under Grant No. 2022010801010273.

\section*{Declarations}

\begin{itemize}

\item Competing interests: The authors declare that they have no known competing financial interests or personal relationships that could have appeared to influence the work reported in this paper.
\item Authors' contributions: \textbf{Yinting Wu}: Conceptualization, Methodology, Formal analysis and investigation, Writing-original draft preparation; \textbf{Peng Pai}: Writing-review and editing; \textbf{Bo Cai}: Writing-review and editing, Funding acquisition; \textbf{Le Li}: Formal analysis and investigation, Supervision, Writing-review and editing, Funding acquisition.
\item Ethical and informed consent for data used: Ethics Committee approval was obtained from the Institutional Ethics Committee of Central China Normal University to the commencement of the study. Moreover, the authors declare that the manuscript is not submitted to more than one journal for simultaneous consideration; The submitted work is original and was not published elsewhere in any form or language (partially or in full); A single study is not be split up into several parts to increase the quantity of submissions and submitted to various journals or to one journal over time; Results are presented clearly, honestly, and without fabrication, falsification or inappropriate data manipulation; No data, text, or theories by others are presented as if they were the author’s own and proper acknowledgements to other works are given.
\item Data availability and access: The data that support the findings of this study are openly available in \href{http://ufldl.stanford.edu/housenumbers/}{SVHN} and \href{https://www.cs.toronto.edu/~kriz/cifar.html}{CIFAR-10/CIFAR-100}
\end{itemize}

\clearpage
\begin{appendices}

\section{Landscapes of models trained under different schemes}\label{secA1}

\vspace{-0.63671875cm}
\begin{figure}[h!]
\hspace{-3.75cm}
\includegraphics[width=1.5\linewidth]{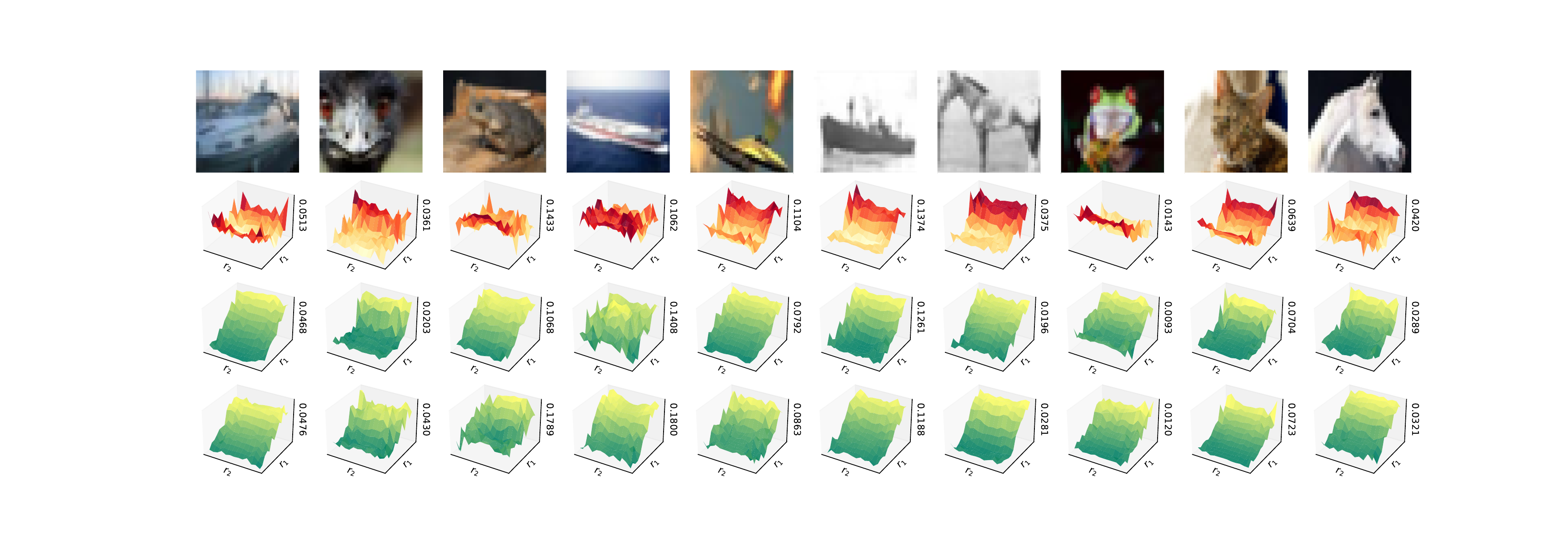}
\vspace{-1.2cm}
\caption{Loss landscapes of three models on ten CIFAR-10 images. Each row corresponds to one of three models trained with plain N-FGSM (first row, red tone), BB(G)-N-FGSM (second row), and BB(BG)-N-FGSM (third row) respectively. Elements in subfigures are same as that of Figure \ref{fig:loss_surface_both}}
\vspace{-0.4cm}
\label{fig:loss_surface_single}
\end{figure}

\vspace{-0.63671875cm}
\begin{figure}[h!]
\hspace{-3.75cm}
\includegraphics[width=1.5\linewidth]{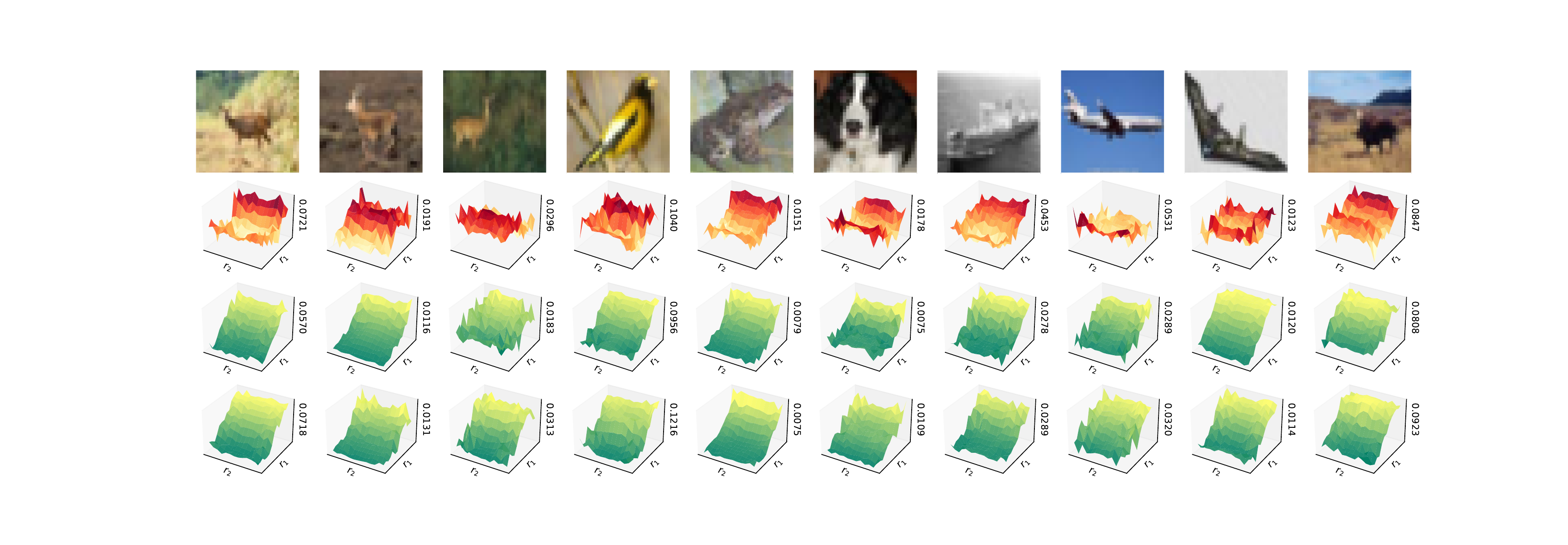}
\vspace{-1.2cm}
\caption{Loss landscapes of three models on ten CIFAR-10 images. Each row corresponds to one of three models trained with plain PGD-10 (first row, red tone), BB(G)-PGD-10 (second row), and BB(BG)-PGD-10 (third row) respectively. The rest is the same as in Figure \ref{fig:loss_surface_both}}
\vspace{-0.4cm}
\label{fig:loss_surface_multi}
\end{figure}

\clearpage
\section{Full results of Table \ref{table:table 1}.}\label{ap:1}

\begin{table}[h]
\centering
\caption{Experiments of PreActResNet18 on SVHN dataset. Settings are described in Section \ref{sec:Experiments}}
\label{table:SVHN}
\begin{tabular}{
    c|m{4.5em}<{\centering}|m{4.5em}<{\centering}|m{4.5em}<{\centering}|m{4.5em}<{\centering}|m{4.5em}<{\centering}}
\toprule
    & 2/255 & 4/255 & 6/255 & 8/255 & 10/255\\
\hline\hline
    \multirow{2}{8em}{N-FGSM} 
    & $96.38_{\pm0.85}$ & $95.12_{\pm0.32}$ 
    & $93.36_{\pm0.32}$ & $91.60_{\pm0.07}$ 
    & $87.50_{\pm0.46}$\\
    & $85.25_{\pm0.72}$ & $70.02_{\pm0.49}$ 
    & $58.36_{\pm0.39}$ 
    & $44.14_{\pm0.61}$ & $32.89_{\pm0.22}$ \\
\hline
    \multirow{2}{8em}{BB(CP)-N-FGSM} 
    & $96.78_{\pm0.62}$ & $95.21_{\pm0.11}$ 
    & $93.29_{\pm0.28}$ 
    & $91.41_{\pm0.86}$ & $88.82_{\pm0.49}$ \\
    & $85.46_{\pm0.89}$ 
    & $70.52_{\pm0.12}$ & $58.48_{\pm0.46}$ 
    & $44.38_{\pm0.67}$ & $33.72_{\pm0.04}$ \\
\hline
    \multirow{2}{8em}{BB(GS)-N-FGSM} 
    & $96.09_{\pm0.85}$ 
    & $94.34_{\pm0.60}$ & $93.73_{\pm0.13}$ 
    & $92.77_{\pm0.27}$ 
    & $89.95_{\pm0.29}$ \\
    & $\bm{87.89_{\pm0.53}}$ & $\bm{75.68_{\pm0.15}}$ 
    & $\bm{62.05_{\pm0.91}}$ 
    & $\bm{51.07_{\pm0.32}}$ & $\bm{35.19_{\pm0.56}}$ \\
\hline
    \multirow{2}{8em}{BB(BG)-N-FGSM} 
    & $95.60_{\pm0.52}$ & $94.24_{\pm0.90}$ 
    & $92.77_{\pm0.78}$ 
    & $91.65_{\pm0.52}$ & $88.00_{\pm0.46}$ \\
    & $84.77_{\pm0.51}$ & $70.75_{\pm1.02}$ 
    & $59.19_{\pm0.06}$ 
    & $46.08_{\pm0.92}$ & $33.42_{\pm0.58}$ \\
\midrule
    \multirow{2}{8em}{PGD-10} & $96.58_{\pm0.45}$ 
    & $95.50_{\pm0.45}$ & $94.07_{\pm0.37}$ 
    & $90.31_{\pm0.43}$ 
    & $85.75_{\pm0.30}$ \\
    & $85.55_{\pm0.40}$ & $71.68_{\pm0.41}$ 
    & $62.18_{\pm0.07}$ 
    & $53.06_{\pm0.41}$ & $43.89_{\pm0.66}$ \\
\hline
    \multirow{2}{8em}{BB(CP)-PGD-10} 
    & $96.48_{\pm0.10}$ & $95.31_{\pm2.06}$ 
    & $94.33_{\pm0.84}$ 
    & $90.98_{\pm0.57}$ & $86.10_{\pm0.84}$ \\
    & $86.91_{\pm0.20}$ & $71.97_{\pm0.34}$ 
    & $62.48_{\pm0.26}$ 
    & $53.52_{\pm0.05}$ & $44.36_{\pm0.29}$ \\
\hline
    \multirow{2}{8em}{BB(GS)-PGD-10} 
    & $96.09_{\pm0.20}$ 
    & $92.77_{\pm0.61}$ & $90.62_{\pm0.68}$ 
    & $88.48_{\pm0.27}$ 
    & $84.94_{\pm0.71}$\\
    & $\bm{87.89_{\pm0.42}}$ & $\bm{79.30_{\pm0.29}}$ 
    & $\bm{66.81_{\pm0.14}}$ 
    & $\bm{62.34_{\pm0.15}}$ & $\bm{50.16_{\pm0.46}}$ \\
\hline
    \multirow{2}{8em}{BB(BG)-PGD-10} 
    & $97.43_{\pm0.72}$ & $96.33_{\pm0.17}$ 
    & $94.29_{\pm0.33}$ 
    & $95.61_{\pm0.46}$ & $86.19_{\pm0.28}$ \\
    & $84.65_{\pm0.84}$ & $70.20_{\pm0.82}$ 
    & $63.12_{\pm0.52}$ 
    & $59.18_{\pm0.71}$ & $49.17_{\pm0.70}$ \\
\bottomrule
\end{tabular}
\end{table}

\begin{sidewaystable}[t]
\caption{Experiments of PreActResNet18 on CIFAR-10 dataset. Settings are described in Section \ref{sec:Experiments}}
\label{table:CIFAR-10}
\begin{tabular}{
    c|m{4.5em}<{\centering}|m{4.5em}<{\centering}|m{4.5em}<{\centering}|m{4.5em}<{\centering}|m{4.5em}<{\centering}|m{4.5em}<{\centering}|m{4.5em}<{\centering}|m{4.5em}<{\centering}}
\toprule
    & 2/255 & 4/255 & 6/255 & 8/255 & 10/255 & 12/255 & 14/255 & 16/255\\
\hline\hline
    \multirow{2}{8em}{N-FGSM} 
    & $90.23_{\pm0.44}$ & $87.58_{\pm0.09}$ 
    & $84.53_{\pm0.47}$ & $80.89_{\pm0.28}$ 
    & $75.76_{\pm0.54}$ 
    & $71.92_{\pm0.52}$ & $66.52_{\pm0.24}$ 
    & $59.53_{\pm0.16}$ \\
    & $77.07_{\pm0.83}$ & $64.44_{\pm0.03}$ 
    & $55.26_{\pm0.11}$ 
    & $44.43_{\pm0.70}$ & $40.86_{\pm1.19}$ 
    & $36.58_{\pm0.22}$ 
    & $32.03_{\pm0.80}$ & $26.16_{\pm0.22}$ \\
\hline
    \multirow{2}{8em}{BB(CP)-N-FGSM} 
    & $91.80_{\pm0.68}$ & $86.91_{\pm0.77}$ 
    & $83.89_{\pm0.22}$ 
    & $79.59_{\pm0.45}$ & $74.40_{\pm0.33}$ 
    & $68.93_{\pm0.99}$ 
    & $62.43_{\pm0.96}$ & $56.93_{\pm0.68}$ \\
    & $78.13_{\pm0.83}$ 
    & $64.58_{\pm0.35}$ & $55.80_{\pm0.88}$ 
    & $45.12_{\pm0.31}$ & $40.88_{\pm0.25}$ 
    & $37.44_{\pm1.72}$ 
    & $32.51_{\pm0.79}$ & $26.66_{\pm0.78}$ \\
\hline
    \multirow{2}{8em}{BB(GS)-N-FGSM} & $89.61_{\pm0.48}$ 
    & $86.94_{\pm0.24}$ & $83.48_{\pm1.28}$ 
    & $84.74_{\pm0.38}$ 
    & $78.16_{\pm0.69}$ & $72.18_{\pm0.11}$ 
    & $66.75_{\pm0.32}$ 
    & $60.70_{\pm1.27}$ \\
    & $\bm{78.32_{\pm0.66}}$ & $\bm{65.62_{\pm0.17}}$ 
    & $\bm{56.48_{\pm0.21}}$ 
    & $\bm{48.24_{\pm0.82}}$ & $\bm{43.06_{\pm0.49}}$ 
    & $\bm{38.98_{\pm0.06}}$ 
    & $\bm{34.35_{\pm0.69}}$ & $\bm{29.14_{\pm0.88}}$ \\
\hline
    \multirow{2}{8em}{BB(BG)-N-FGSM} 
    & $87.66_{\pm0.16}$ & $86.51_{\pm0.24}$ 
    & $84.90_{\pm0.76}$ 
    & $83.50_{\pm0.03}$ & $78.39_{\pm0.37}$ 
    & $72.92_{\pm0.50}$ 
    & $67.45_{\pm0.18}$ & $61.13_{\pm0.76}$ \\
    & $73.28_{\pm0.57}$ & $61.56_{\pm0.75}$ 
    & $53.51_{\pm0.12}$ 
    & $45.51_{\pm0.46}$ & $40.04_{\pm0.61}$ 
    & $35.04_{\pm0.23}$ 
    & $30.13_{\pm0.49}$ & $25.82_{\pm0.73}$ \\
\midrule
    \multirow{2}{8em}{PGD-10} & $90.86_{\pm0.44}$ 
    & $86.09_{\pm0.42}$ & $85.14_{\pm0.83}$ 
    & $84.38_{\pm0.18}$ 
    & $78.49_{\pm0.20}$ & $71.44_{\pm0.18}$ 
    & $65.40_{\pm0.38}$ 
    & $56.87_{\pm0.11}$ \\
    & $77.34_{\pm0.93}$ & $66.45_{\pm0.06}$ 
    & $57.21_{\pm0.34}$ 
    & $48.44_{\pm0.16}$ & $45.30_{\pm0.73}$ 
    & $40.32_{\pm0.60}$ 
    & $36.81_{\pm0.69}$ & $30.70_{\pm0.67}$ \\
\hline
    \multirow{2}{8em}{BB(CP)-PGD-10} 
    & $91.02_{\pm0.15}$ & $88.48_{\pm1.28}$ 
    & $84.74_{\pm0.63}$ 
    & $80.86_{\pm0.27}$ & $75.70_{\pm0.58}$ 
    & $69.47_{\pm0.05}$ 
    & $63.74_{\pm0.35}$ & $57.62_{\pm0.22}$ \\
    & $\bm{79.00_{\pm0.26}}$ & $66.34_{\pm0.39}$ 
    & $57.54_{\pm0.77}$ 
    & $48.85_{\pm0.32}$ & $45.98_{\pm2.31}$ 
    & $40.85_{\pm0.64}$ 
    & $\bm{37.46_{\pm0.79}}$ & $31.64_{\pm0.28}$ \\
\hline
    \multirow{2}{8em}{BB(GS)-PGD-10} & $90.57_{\pm0.89}$ 
    & $85.20_{\pm0.53}$ & $83.07_{\pm0.93}$ & $81.25_{\pm0.59}$ 
    & $76.84_{\pm0.26}$ & $71.31_{\pm0.23}$ 
    & $67.20_{\pm0.11}$ 
    & $62.32_{\pm0.89}$ \\
    & $77.65_{\pm0.60}$ & $\bm{67.08_{\pm0.35}}$ 
    & $\bm{59.58_{\pm0.20}}$ 
    & $\bm{51.76_{\pm1.54}}$ & $\bm{46.40_{\pm0.43}}$ 
    & $\bm{42.11_{\pm0.53}}$ 
    & $37.27_{\pm0.26}$ & $\bm{33.36_{\pm0.84}}$ \\
\hline
    \multirow{2}{8em}{BB(BG)-PGD-10} 
    & $91.52_{\pm0.74}$ & $87.97_{\pm0.77}$ 
    & $84.21_{\pm0.65}$ 
    & $83.69_{\pm0.11}$ & $78.33_{\pm0.59}$ 
    & $73.01_{\pm0.12}$ 
    & $67.98_{\pm0.58}$ & $62.97_{\pm0.66}$ \\
    & $74.38_{\pm0.36}$ & $61.64_{\pm0.23}$ 
    & $55.01_{\pm0.68}$ 
    & $49.71_{\pm0.63}$ & $45.44_{\pm0.29}$ 
    & $40.81_{\pm0.48}$ 
    & $36.89_{\pm0.38}$ & $32.11_{\pm0.45}$ \\
\bottomrule
\end{tabular}
\end{sidewaystable}

\begin{sidewaystable}[t]
\centering
\caption{Experiments of PreActResNet18 on CIFAR-100 dataset. Settings are described in Section \ref{sec:Experiments}}
\label{table:CIFAR-100}
\begin{tabular}{
    c|m{4.5em}<{\centering}|m{4.5em}<{\centering}|m{4.5em}<{\centering}|m{4.5em}<{\centering}|m{4.5em}<{\centering}|m{4.5em}<{\centering}|m{4.5em}<{\centering}|m{4.5em}<{\centering}}
\toprule
    & 2/255 & 4/255 & 6/255 & 8/255 & 10/255 & 12/255 & 14/255 & 16/255\\
\hline\hline
    \multirow{2}{8em}{N-FGSM} 
    & $64.94_{\pm0.07}$ & $60.06_{\pm0.69}$ 
    & $55.60_{\pm0.31}$ & $50.78_{\pm0.11}$ 
    & $46.06_{\pm0.16}$ 
    & $42.16_{\pm0.77}$ & $38.00_{\pm0.68}$ 
    & $34.86_{\pm0.13}$ \\
    & $47.66_{\pm0.62}$ & $36.81_{\pm0.85}$ 
    & $29.68_{\pm0.73}$ 
    & $23.14_{\pm0.10}$ & $20.12_{\pm0.55}$ 
    & $18.07_{\pm0.69}$ 
    & $16.27_{\pm0.60}$ & $14.45_{\pm0.12}$ \\
\hline
    \multirow{2}{8em}{BB(CP)-N-FGSM} 
    & $66.31_{\pm0.20}$ & $60.94_{\pm0.35}$ 
    & $56.36_{\pm0.30}$ 
    & $52.25_{\pm0.23}$ & $46.15_{\pm0.88}$ 
    & $41.45_{\pm0.18}$ 
    & $39.84_{\pm0.50}$ & $37.01_{\pm0.08}$ \\
    & $\bm{48.54_{\pm0.17}}$ 
    & $\bm{36.93_{\pm0.44}}$ & $\bm{31.99_{\pm0.64}}$ 
    & $\bm{25.39_{\pm0.20}}$ & $\bm{22.22_{\pm0.17}}$ 
    & $\bm{19.09_{\pm0.44}}$ 
    & $\bm{16.98_{\pm0.26}}$ & $\bm{14.55_{\pm0.24}}$ \\
\hline
    \multirow{2}{8em}{BB(GS)-N-FGSM} 
    & $62.60_{\pm0.54}$ 
    & $58.77_{\pm0.63}$ & $53.43_{\pm0.31}$ 
    & $48.14_{\pm0.68}$ 
    & $43.47_{\pm0.35}$ & $39.78_{\pm0.69}$ 
    & $35.86_{\pm0.54}$ 
    & $31.34_{\pm0.50}$ \\
    & $45.90_{\pm0.39}$ & $34.86_{\pm0.51}$ 
    & $28.54_{\pm0.30}$ 
    & $23.74_{\pm0.24}$ & $18.47_{\pm0.75}$ 
    & $15.70_{\pm0.60}$ 
    & $12.55_{\pm0.30}$ & $11.40_{\pm0.30}$ \\
\hline
    \multirow{2}{8em}{BB(BG)-N-FGSM} 
    & $63.77_{\pm0.64}$ & $61.17_{\pm0.69}$ 
    & $56.81_{\pm0.14}$ 
    & $51.44_{\pm0.57}$ & $46.99_{\pm0.32}$ 
    & $41.36_{\pm0.62}$ 
    & $36.85_{\pm0.47}$ & $31.64_{\pm0.59}$ \\
    & $44.82_{\pm0.59}$ & $33.59_{\pm0.38}$ 
    & $27.87_{\pm0.23}$ 
    & $22.31_{\pm0.23}$ & $18.56_{\pm0.29}$ 
    & $15.07_{\pm0.46}$ 
    & $12.84_{\pm0.36}$ & $10.64_{\pm0.04}$ \\
\midrule
    \multirow{2}{8em}{PGD-10} 
    & $66.01_{\pm0.64}$ 
    & $61.33_{\pm0.52}$ & $57.20_{\pm0.08}$ 
    & $54.59_{\pm0.96}$ 
    & $50.85_{\pm0.91}$ & $45.55_{\pm1.18}$ 
    & $40.49_{\pm0.35}$ 
    & $36.13_{\pm0.39}$ \\
    & $47.95_{\pm0.67}$ & $37.79_{\pm0.84}$ 
    & $30.44_{\pm0.10}$ 
    & $25.59_{\pm0.02}$ & $22.33_{\pm0.37}$ 
    & $20.82_{\pm0.57}$ 
    & $18.11_{\pm0.56}$ & $16.55_{\pm0.23}$ \\
\hline
    \multirow{2}{8em}{BB(CP)-PGD-10} 
    & $66.21_{\pm0.57}$ & $61.72_{\pm0.48}$ 
    & $56.64_{\pm0.20}$ 
    & $53.16_{\pm0.13}$ & $48.83_{\pm0.73}$ 
    & $44.69_{\pm0.41}$ 
    & $39.31_{\pm0.87}$ & $35.94_{\pm0.69}$ \\
    & $\bm{48.08_{\pm0.17}}$ & $\bm{38.51_{\pm0.64}}$ 
    & $\bm{31.20_{\pm0.30}}$ 
    & $\bm{28.32_{\pm0.35}}$ & $\bm{23.01_{\pm1.02}}$ 
    & $\bm{21.26_{\pm0.70}}$ 
    & $\bm{18.82_{\pm0.16}}$ & $\bm{17.41_{\pm0.10}}$ \\
\hline
    \multirow{2}{8em}{BB(GS)-PGD-10} 
    & $63.67_{\pm0.56}$ 
    & $58.89_{\pm0.45}$ & $52.26_{\pm0.14}$ 
    & $46.68_{\pm0.08}$ 
    & $43.56_{\pm0.36}$ & $40.58_{\pm0.90}$ 
    & $37.57_{\pm0.71}$ 
    & $33.79_{\pm0.05}$ \\
    & $46.54_{\pm0.73}$ & $35.59_{\pm0.52}$ 
    & $29.79_{\pm0.81}$ 
    & $23.50_{\pm0.41}$ & $20.43_{\pm0.22}$ 
    & $19.52_{\pm0.77}$ 
    & $17.86_{\pm0.42}$ & $15.82_{\pm0.32}$ \\
\hline
    \multirow{2}{8em}{BB(BG)-PGD-10} 
    & $64.45_{\pm0.86}$ & $62.02_{\pm0.17}$ 
    & $59.77_{\pm0.39}$ 
    & $56.63_{\pm0.05}$ & $52.12_{\pm0.64}$ 
    & $48.88_{\pm0.18}$ 
    & $44.64_{\pm0.94}$ & $40.39_{\pm0.15}$ \\
    & $45.12_{\pm0.32}$ & $33.50_{\pm0.54}$ 
    & $28.16_{\pm0.53}$ 
    & $22.51_{\pm0.52}$ & $20.74_{\pm0.65}$ 
    & $18.99_{\pm0.53}$ 
    & $16.13_{\pm0.86}$ & $14.16_{\pm0.17}$ \\
\bottomrule
\end{tabular}
\end{sidewaystable}

\begin{table}[t]
\centering
\caption{Experiments of WideResNet28-10 on SVHN dataset. Settings are described in Section \ref{sec:Experiments}}
\label{table:WRN28,SVHN}
\begin{tabular}{
    c|m{4.5em}<{\centering}|m{4.5em}<{\centering}|m{4.5em}<{\centering}|m{4.5em}<{\centering}|m{4.5em}<{\centering}}
\toprule
    & 2/255 & 4/255 & 6/255 & 8/255 & 10/255\\
\hline\hline
    \multirow{2}{8em}{N-FGSM} 
    & $96.80_{\pm0.49}$ & $95.39_{\pm0.82}$ 
    & $93.30_{\pm1.08}$ & $91.48_{\pm0.13}$ 
    & $86.25_{\pm1.04}$\\
    & $86.33_{\pm0.84}$ & $71.17_{\pm0.67}$ 
    & $56.48_{\pm0.34}$ 
    & $42.03_{\pm0.79}$ & $31.56_{\pm0.40}$ \\
\hline
    \multirow{2}{8em}{BB(CP)-N-FGSM} 
    & $97.17_{\pm0.88}$ & $95.31_{\pm0.53}$ 
    & $94.98_{\pm0.33}$ 
    & $89.84_{\pm0.43}$ & $86.72_{\pm0.52}$ \\
    & $86.13_{\pm0.69}$ 
    & $70.21_{\pm0.93}$ & $55.89_{\pm0.65}$ 
    & $42.56_{\pm0.19}$ & $31.25_{\pm0.41}$ \\
\hline
    \multirow{2}{8em}{BB(GS)-N-FGSM} 
    & $95.41_{\pm0.06}$ 
    & $93.95_{\pm0.10}$ & $91.73_{\pm0.78}$ 
    & $90.16_{\pm0.96}$ 
    & $89.36_{\pm0.11}$ \\
    & \bm{$89.65_{\pm0.73}$} & $78.61_{\pm0.47}$ 
    & $\bm{66.49_{\pm0.71}}$ 
    & $\bm{55.37_{\pm0.07}}$ & $\bm{39.65_{\pm0.86}}$ \\
\hline
    \multirow{2}{8em}{BB(BG)-N-FGSM} 
    & $96.33_{\pm0.68}$ & $94.53_{\pm0.85}$ 
    & $94.22_{\pm0.99}$ 
    & $94.37_{\pm0.55}$ & $85.55_{\pm0.27}$ \\
    & $88.67_{\pm0.63}$ & \bm{$79.10_{\pm0.44}$} 
    & $61.71_{\pm0.97}$ 
    & $51.48_{\pm0.90}$ & $28.32_{\pm0.89}$ \\
\midrule
    \multirow{2}{8em}{PGD-10}
    & $96.87_{\pm0.54}$ 
    & $95.62_{\pm0.50}$ & $94.79_{\pm0.61}$ 
    & $92.58_{\pm0.30}$ 
    & $89.14_{\pm0.76}$ \\
    & $87.58_{\pm0.60}$ & $73.44_{\pm0.20}$ 
    & $62.34_{\pm0.57}$ 
    & $51.64_{\pm0.22}$ & $42.19_{\pm0.48}$ \\
\hline
    \multirow{2}{8em}{BB(CP)-PGD-10} 
    & $96.72_{\pm0.66}$ & $96.09_{\pm0.17}$ 
    & $93.23_{\pm0.64}$ 
    & $91.72_{\pm0.62}$ & $88.59_{\pm0.87}$ \\
    & $87.19_{\pm0.12}$ & $73.83_{\pm0.81}$ 
    & $62.56_{\pm0.95}$ 
    & $50.78_{\pm0.23}$ & $42.42_{\pm0.80}$ \\
\hline
    \multirow{2}{8em}{BB(GS)-PGD-10} 
    & $94.30_{\pm0.83}$ 
    & $93.59_{\pm0.35}$ & $92.29_{\pm0.51}$ 
    & $91.33_{\pm0.05}$ 
    & $86.33_{\pm0.36}$\\
    & $\bm{90.16_{\pm1.09}}$ & $\bm{83.20_{\pm0.28}}$ 
    & $\bm{71.43_{\pm0.70}}$ 
    & $59.38_{\pm0.15}$ & $45.63_{\pm0.42}$ \\
\hline
    \multirow{2}{8em}{BB(BG)-PGD-10} 
    & $94.77_{\pm0.45}$ & $94.37_{\pm0.56}$ 
    & $92.93_{\pm0.29}$ 
    & $91.61_{\pm0.39}$ & $89.84_{\pm0.14}$ \\
    & $89.45_{\pm0.92}$ & $81.25_{\pm0.21}$ 
    & $70.40_{\pm1.07}$ 
    & $\bm{60.08_{\pm0.31}}$ & $\bm{47.81_{\pm0.25}}$ \\
\bottomrule
\end{tabular}
\end{table}

\begin{sidewaystable}[t]
\centering
\caption{Experiments of WideResNet28-10 on CIFAR-10 dataset. Settings are described in Section \ref{sec:Experiments}}
\label{table:WRN28,CIFAR-10}
\begin{tabular}{
    c|m{4.5em}<{\centering}|m{4.5em}<{\centering}|m{4.5em}<{\centering}|m{4.5em}<{\centering}|m{4.5em}<{\centering}|m{4.5em}<{\centering}|m{4.5em}<{\centering}|m{4.5em}<{\centering}}
\toprule
    & 2/255 & 4/255 & 6/255 & 8/255 & 10/255 & 12/255 & 14/255 & 16/255\\
\hline\hline
    \multirow{2}{8em}{N-FGSM} 
    & $90.23_{\pm0.10}$ & $85.06_{\pm0.36}$ 
    & $81.17_{\pm0.96}$ & $77.15_{\pm2.02}$ 
    & $72.87_{\pm0.63}$ 
    & $68.67_{\pm0.62}$ & $63.50_{\pm0.92}$ 
    & $59.28_{\pm0.65}$ \\
    & $77.93_{\pm0.71}$ & $63.67_{\pm0.12}$ 
    & $54.37_{\pm0.44}$ 
    & $44.82_{\pm0.19}$ & $40.28_{\pm0.17}$ 
    & $35.22_{\pm0.15}$ 
    & $30.41_{\pm0.61}$ & $\bm{26.17_{\pm0.76}}$ \\
\hline
    \multirow{2}{8em}{BB(CP)-N-FGSM} 
    & $91.33_{\pm0.81}$ & $87.19_{\pm1.30}$ 
    & $83.82_{\pm0.70}$ 
    & $79.22_{\pm0.27}$ & $74.04_{\pm0.54}$ 
    & $69.28_{\pm0.28}$ 
    & $64.73_{\pm0.23}$ & $58.83_{\pm0.80}$ \\
    & $\bm{78.05_{\pm0.74}}$ 
    & $\bm{64.53_{\pm0.32}}$ & $54.47_{\pm0.68}$ 
    & $43.91_{\pm0.72}$ & $39.35_{\pm0.94}$ 
    & $34.88_{\pm0.86}$ 
    & $29.97_{\pm0.33}$ & $24.45_{\pm0.91}$ \\
\hline
    \multirow{2}{8em}{BB(GS)-N-FGSM}
    & $88.44_{\pm0.46}$ 
    & $84.69_{\pm0.26}$ & $80.61_{\pm0.45}$ 
    & $76.09_{\pm0.05}$ 
    & $72.76_{\pm0.95}$ & $69.61_{\pm0.25}$ 
    & $65.57_{\pm0.79}$ 
    & $62.27_{\pm0.87}$ \\
    & $74.30_{\pm0.89}$ & $63.67_{\pm0.85}$ 
    & $\bm{55.81_{\pm0.59}}$ 
    & $\bm{47.11_{\pm0.21}}$ & $\bm{42.52_{\pm0.78}}$ 
    & $\bm{36.90_{\pm0.57}}$ 
    & $\bm{31.62_{\pm0.11}}$ & $26.02_{\pm0.64}$ \\
\hline
    \multirow{2}{8em}{BB(BG)-N-FGSM} 
    & $89.16_{\pm0.20}$ & $85.23_{\pm0.46}$ 
    & $82.32_{\pm0.14}$ 
    & $79.14_{\pm0.38}$ & $73.82_{\pm0.39}$ 
    & $68.60_{\pm0.49}$ 
    & $62.29_{\pm1.08}$ & $57.19_{\pm0.07}$ \\
    & $74.02_{\pm0.47}$ & $62.97_{\pm0.41}$ 
    & $54.76_{\pm0.34}$ 
    & $46.72_{\pm0.13}$ & $40.98_{\pm0.93}$ 
    & $35.10_{\pm0.53}$ 
    & $29.40_{\pm0.84}$ & $23.28_{\pm0.06}$ \\
\midrule
    \multirow{2}{8em}{PGD-10}
    & $90.62_{\pm0.90}$ 
    & $87.66_{\pm0.56}$ & $82.79_{\pm0.48}$ 
    & $77.89_{\pm0.97}$ 
    & $73.43_{\pm0.99}$ & $68.46_{\pm0.52}$ 
    & $63.39_{\pm0.66}$ 
    & $58.67_{\pm0.18}$ \\
    & $\bm{77.27_{\pm0.51}}$ & $65.39_{\pm0.40}$ 
    & $55.44_{\pm0.58}$ 
    & $46.33_{\pm0.50}$ & $41.31_{\pm0.55}$ 
    & $36.21_{\pm0.67}$ 
    & $31.58_{\pm0.24}$ & $27.42_{\pm0.35}$ \\
\hline
    \multirow{2}{8em}{BB(CP)-PGD-10} 
    & $90.33_{\pm0.42}$ & $87.50_{\pm0.37}$ 
    & $83.13_{\pm0.83}$ 
    & $78.61_{\pm1.04}$ & $73.61_{\pm0.43}$ 
    & $68.50_{\pm0.09}$ 
    & $63.79_{\pm0.88}$ & $58.89_{\pm0.31}$ \\
    & $77.15_{\pm0.01}$ & $\bm{65.43_{\pm0.75}}$ 
    & $55.68_{\pm0.73}$ 
    & $46.39_{\pm1.16}$ & $42.44_{\pm0.60}$ 
    & $37.94_{\pm0.22}$ 
    & $33.93_{\pm0.03}$ & $29.10_{\pm0.29}$ \\
\hline
    \multirow{2}{8em}{BB(GS)-PGD-10}
    & $88.87_{\pm0.77}$ 
    & $84.57_{\pm0.98}$ & $80.73_{\pm0.69}$ & $76.56_{\pm0.82}$ 
    & $72.50_{\pm0.60}$ & $68.71_{\pm0.76}$ 
    & $65.58_{\pm0.28}$ 
    & $61.33_{\pm0.51}$ \\
    & $76.07_{\pm0.85}$ & $63.67_{\pm0.39}$ 
    & $\bm{56.87_{\pm0.97}}$ 
    & $\bm{49.61_{\pm0.96}}$ & $\bm{44.93_{\pm0.33}}$ 
    & $\bm{40.04_{\pm0.32}}$ 
    & $\bm{35.75_{\pm0.81}}$ & $\bm{30.76_{\pm0.54}}$ \\
\hline
    \multirow{2}{8em}{BB(BG)-PGD-10} 
    & $88.28_{\pm0.44}$ & $85.45_{\pm0.29}$ 
    & $81.92_{\pm0.20}$ 
    & $77.54_{\pm0.67}$ & $74.70_{\pm0.56}$ 
    & $70.94_{\pm0.62}$ 
    & $67.41_{\pm0.66}$ & $63.87_{\pm0.89}$ \\
    & $75.79_{\pm0.11}$ & $64.45_{\pm0.26}$ 
    & $56.21_{\pm0.69}$ 
    & $48.34_{\pm0.42}$ & $43.23_{\pm0.65}$ 
    & $38.54_{\pm0.22}$ 
    & $33.85_{\pm0.34}$ & $29.30_{\pm0.19}$ \\
\bottomrule
\end{tabular}
\end{sidewaystable}

\begin{sidewaystable}[t]
\centering
\caption{Experiments of WideResNet28-10 on CIFAR-100 dataset. Settings are described in Section \ref{sec:Experiments}}
\label{table:WRN28,CIFAR-100}
\begin{tabular}{
    c|m{4.5em}<{\centering}|m{4.5em}<{\centering}|m{4.5em}<{\centering}|m{4.5em}<{\centering}|m{4.5em}<{\centering}|m{4.5em}<{\centering}|m{4.5em}<{\centering}|m{4.5em}<{\centering}}
\toprule
    & 2/255 & 4/255 & 6/255 & 8/255 & 10/255 & 12/255 & 14/255 & 16/255\\
\hline\hline
    \multirow{2}{8em}{N-FGSM} 
    & $68.28_{\pm0.97}$ & $61.80_{\pm0.56}$ 
    & $55.04_{\pm0.61}$ & $48.75_{\pm0.65}$ 
    & $44.10_{\pm0.68}$ 
    & $39.32_{\pm1.34}$ & $34.63_{\pm0.60}$ 
    & $30.08_{\pm0.42}$ \\
    & $\bm{49.84_{\pm0.95}}$ & $36.09_{\pm0.75}$ 
    & $29.81_{\pm0.38}$ 
    & $23.75_{\pm0.45}$ & $21.95_{\pm0.40}$ 
    & $18.65_{\pm0.18}$ 
    & $15.94_{\pm0.13}$ & $12.81_{\pm0.87}$ \\
\hline
    \multirow{2}{8em}{BB(CP)-N-FGSM} 
    & $66.89_{\pm0.53}$ & $61.04_{\pm0.76}$ 
    & $55.46_{\pm0.36}$ 
    & $49.02_{\pm0.66}$ & $45.91_{\pm0.83}$ 
    & $41.87_{\pm0.11}$ 
    & $38.73_{\pm0.77}$ & $34.57_{\pm0.74}$ \\
    & $49.80_{\pm0.31}$ 
    & $\bm{37.01_{\pm0.62}}$ & $\bm{30.90_{\pm0.58}}$ 
    & $\bm{24.32_{\pm0.72}}$ & $\bm{22.70_{\pm2.05}}$ 
    & $\bm{19.77_{\pm0.20}}$ 
    & $\bm{16.30_{\pm0.23}}$ & $\bm{13.48_{\pm0.48}}$ \\
\hline
    \multirow{2}{8em}{BB(GS)-N-FGSM} 
    & $65.14_{\pm0.06}$ & $59.57_{\pm0.90}$ 
    & $55.28_{\pm0.04}$ 
    & $51.66_{\pm0.50}$ & $47.79_{\pm0.71}$ 
    & $42.75_{\pm0.21}$ 
    & $38.22_{\pm0.24}$ & $33.79_{\pm0.80}$ \\
    & $47.07_{\pm0.54}$ & $35.25_{\pm0.82}$ 
    & $29.90_{\pm0.94}$ 
    & $23.14_{\pm0.81}$ & $20.32_{\pm0.67}$ 
    & $17.32_{\pm0.43}$ 
    & $15.78_{\pm0.98}$ & $12.70_{\pm0.30}$ \\
\hline
    \multirow{2}{8em}{BB(BG)-N-FGSM} 
    & $65.92_{\pm0.46}$ 
    & $60.64_{\pm0.35}$ & $57.65_{\pm0.33}$ 
    & $53.71_{\pm0.78}$ 
    & $47.31_{\pm0.41}$ & $41.65_{\pm0.15}$ 
    & $35.16_{\pm0.14}$ 
    & $29.88_{\pm0.03}$ \\
    & $46.29_{\pm0.69}$ & $33.59_{\pm0.32}$ 
    & $27.28_{\pm0.55}$ 
    & $21.39_{\pm0.86}$ & $18.19_{\pm0.85}$ 
    & $15.87_{\pm0.70}$ 
    & $12.63_{\pm0.11}$ & $8.89_{\pm1.10}$ \\
\midrule
    \multirow{2}{8em}{PGD-10} 
    & $67.42_{\pm0.22}$ 
    & $60.78_{\pm0.97}$ & $56.54_{\pm0.74}$ 
    & $51.48_{\pm0.55}$ 
    & $46.17_{\pm0.40}$ & $42.37_{\pm0.35}$ 
    & $37.72_{\pm0.03}$ 
    & $32.89_{\pm0.38}$ \\
    & $49.84_{\pm0.93}$ & ${36.88_{\pm0.81}}$ 
    & $31.78_{\pm0.99}$ 
    & $25.23_{\pm0.69}$ & $22.05_{\pm0.82}$ 
    & $19.84_{\pm0.39}$ 
    & $16.59_{\pm0.61}$ & $13.75_{\pm0.12}$ \\
\hline
    \multirow{2}{8em}{BB(CP)-PGD-10} 
    & $68.20_{\pm0.23}$ & $61.25_{\pm0.21}$ 
    & $55.72_{\pm0.68}$ 
    & $50.47_{\pm0.44}$ & $45.47_{\pm0.29}$ 
    & $41.29_{\pm0.47}$ 
    & $36.85_{\pm0.53}$ & $32.66_{\pm0.57}$ \\
    & $\bm{50.39_{\pm0.36}}$ & $\bm{37.40_{\pm0.75}}$ 
    & $\bm{32.33_{\pm0.63}}$ 
    & $\bm{25.76_{\pm0.25}}$ & $\bm{22.46_{\pm1.08}}$ 
    & $\bm{20.34_{\pm0.14}}$ 
    & $\bm{17.40_{\pm0.94}}$ & $\bm{13.98_{\pm0.71}}$ \\
\hline
    \multirow{2}{8em}{BB(GS)-PGD-10} 
    & $67.03_{\pm0.52}$ 
    & $61.02_{\pm0.72}$ & $55.50_{\pm0.76}$ 
    & $50.31_{\pm0.37}$ 
    & $46.40_{\pm0.98}$ & $41.81_{\pm0.65}$ 
    & $36.23_{\pm0.33}$ 
    & $31.80_{\pm0.91}$ \\
    & $46.48_{\pm0.73}$ & $35.00_{\pm0.54}$ 
    & $29.72_{\pm0.90}$ 
    & $23.44_{\pm0.46}$ & $20.60_{\pm0.02}$ 
    & $18.12_{\pm0.58}$ 
    & $15.15_{\pm0.50}$ & $13.52_{\pm0.06}$ \\
\hline
    \multirow{2}{8em}{BB(BG)-PGD-10} 
    & $66.95_{\pm0.26}$ & $61.88_{\pm0.04}$ 
    & $56.16_{\pm0.45}$ 
    & $52.03_{\pm0.89}$ & $48.74_{\pm0.92}$ 
    & $45.18_{\pm0.07}$ 
    & $45.08_{\pm0.64}$ & $38.36_{\pm0.83}$ \\
    & $45.94_{\pm0.13}$ & $34.61_{\pm0.83}$ 
    & $28.10_{\pm0.67}$ 
    & $22.27_{\pm0.86}$ & $19.55_{\pm0.56}$ 
    & $17.58_{\pm1.15}$ 
    & $14.34_{\pm0.41}$ & $12.27_{\pm0.48}$ \\
\bottomrule
\end{tabular}
\end{sidewaystable}

\end{appendices}

\clearpage
\bibliography{sn-bibliography}

\end{document}